\documentclass[10pt,twocolumn,letterpaper]{article}

\usepackage[pagenumbers]{cvpr} 

\usepackage[dvipsnames,table,xcdraw]{xcolor}

\usepackage{algorithm}
\usepackage{algorithmic}

\usepackage{amsmath}

\usepackage{multirow}
\usepackage{graphicx}
\usepackage{makecell}
\usepackage{bbding}

\definecolor{tabfirst}{rgb}{1, 0.7, 0.7} 
\definecolor{tabsecond}{rgb}{1, 0.85, 0.7} 
\definecolor{tabthird}{rgb}{1, 1, 0.7} 

\definecolor{cvprblue}{rgb}{0.21,0.49,0.74}
\usepackage[hypcap=false]{caption}
\usepackage[pagebackref,breaklinks,colorlinks,citecolor=cvprblue]{hyperref}
\usepackage{cuted}
\usepackage{multirow}
\usepackage[accsupp]{axessibility} 

\def\paperID{12350} 
\def\confName{CVPR}
\def\confYear{2025}

\title{AnyDressing: Customizable Multi-Garment Virtual Dressing via Latent Diffusion Models}

\author{
    Xinghui Li\textsuperscript{1}
    \quad Qichao Sun\textsuperscript{1}
    \quad\enspace Pengze Zhang\textsuperscript{1}
    \quad Fulong Ye\textsuperscript{1} 
    \quad\enspace Zhichao Liao\textsuperscript{2} \\
    Wanquan Feng\textsuperscript{1\dag}
    \quad\enspace Songtao Zhao\textsuperscript{1\dag}
    \quad Qian He\textsuperscript{1}
    \\\textsuperscript{1}ByteDance \quad \textsuperscript{2}Tsinghua University \\
    \url{https://crayon-shinchan.github.io/AnyDressing/}
}

\begin{document}
\maketitle

{\let\thefootnote\relax\footnote{{\textsuperscript{\dag}Corresponding author.}}}

\begin{strip}
    \vspace*{-15mm}
    \centering
    \includegraphics[width=1\textwidth]{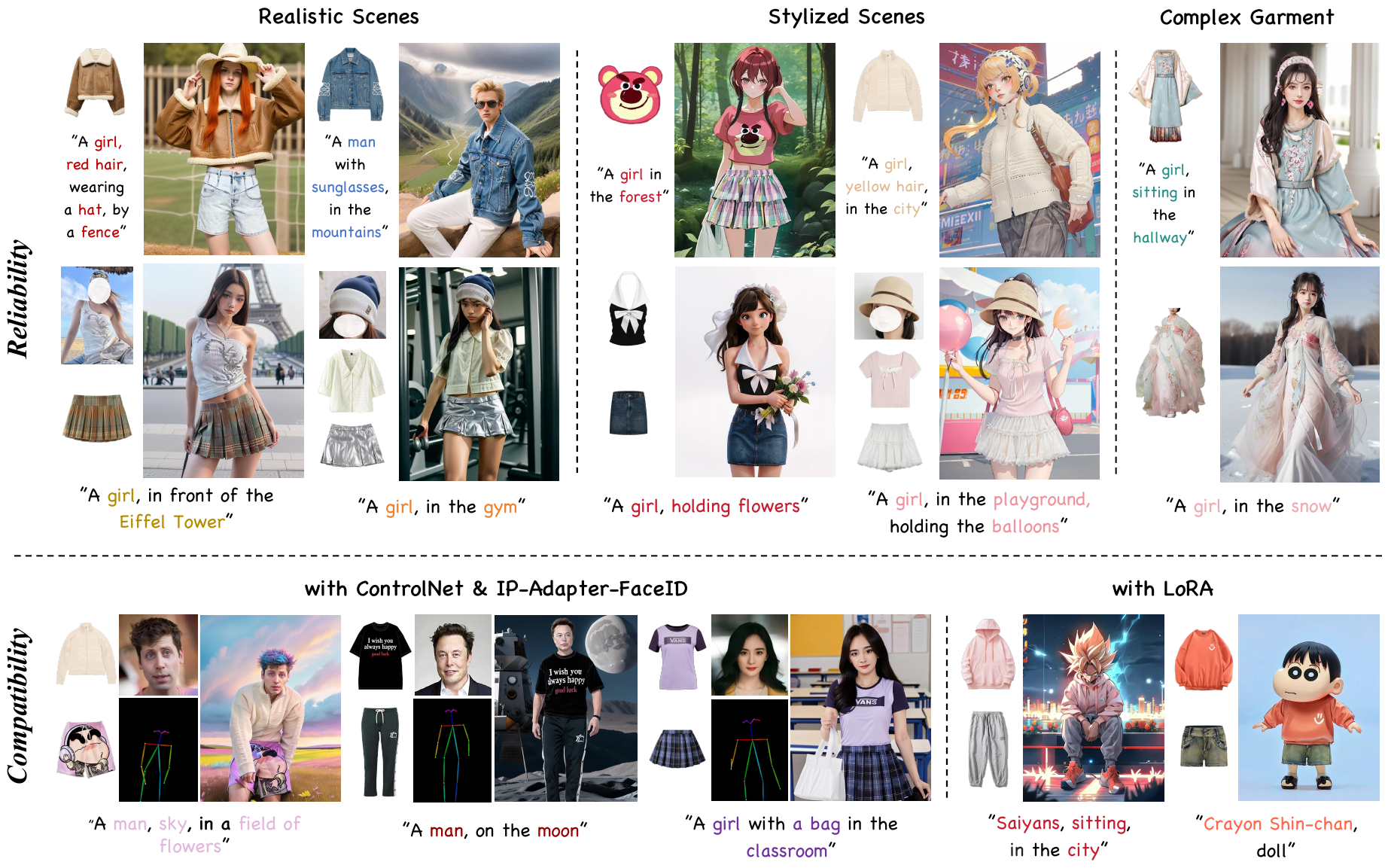}
    \captionof{figure}{\textbf{Customizable virtual dressing results} of our AnyDressing. \textbf{Reliability:} AnyDressing is well-suited for a variety of scenes and complex garments. \textbf{Compatibility:} AnyDressing is compatible with LoRA~\cite{hu2021lora} and plugins such as ControlNet~\cite{zhang2023adding} and FaceID~\cite{ye2023ip}.}
    \label{teaser}
    \vspace*{-2mm}
\end{strip}
\begin{abstract}
Recent advances in garment-centric image generation from text and image prompts based on diffusion models are impressive. However, existing methods lack support for various combinations of attire, and struggle to preserve the garment details while maintaining faithfulness to the text prompts, limiting their performance across diverse scenarios. 
In this paper, we focus on a new task, i.e., \textbf{Multi-Garment Virtual Dressing}, and we propose a novel \textbf{AnyDressing} method for customizing characters conditioned on any combination of garments and any personalized text prompts. 
AnyDressing comprises two primary networks named GarmentsNet and DressingNet, which are respectively dedicated to extracting detailed clothing features and generating customized images.
Specifically, we propose an efficient and scalable module called Garment-Specific Feature Extractor in GarmentsNet to individually encode garment textures in parallel. This design prevents garment confusion while ensuring network efficiency. 
Meanwhile, we design an adaptive Dressing-Attention mechanism and a novel Instance-Level Garment Localization Learning strategy in DressingNet to accurately inject multi-garment features into their corresponding regions. This approach efficiently integrates multi-garment texture cues into generated images and further enhances text-image consistency. 
Additionally, we introduce a Garment-Enhanced Texture Learning strategy to improve the fine-grained texture details of garments. 
Thanks to our well-craft design, AnyDressing can serve as a plug-in module to easily integrate with any community control extensions for diffusion models, improving the diversity and controllability of synthesized images. 
Extensive experiments show that AnyDressing achieves state-of-the-art results. 
\end{abstract}    
\section{Introduction}
In recent years, the field of image generation has experienced transformative advancements~\cite{chang2023muse, ding2021cogview, kang2023scaling}, particularly with methods based on Latent Diffusion Models (LDMs) achieving remarkable success in text-to-image generation tasks~\cite{ho2020denoising, ramesh2022hierarchical, sohl2015deep, song2020score, rombach2022high}. Considering only textual information is inadequate for image customization, numerous approaches incorporate reference images with textual descriptions for image generation~\cite{kumari2023multi, ruiz2023dreambooth, wei2023elite}. Specially, the Virtual Dressing (VD) task of generating garment-centric human images based on the reference garments has sparked considerable research interest~\cite{chen2024magic, wang2024stablegarment, shen2024imagdressing}, due to its significant potential for practical applications in e-commerce and creative design. 


VD is used to be regarded as a subtask of traditional subject-driven image customization, prior approaches~\cite{gal2022image, huang2023reversion, kumari2023multi, liu2023cones, ruiz2023dreambooth, shi2024instantbooth, vinker2023concept, zhang2023sine} simply integrate the features of reference image into the text embeddings without fully exploiting the information from the reference image. Several subsequent works~\cite{ye2023ip, ma2024subject} more comprehensively utilize the features of the reference image by training additional cross-attention layers to integrate reference image features into the diffusion model. However, these methods struggle to preserve the intricate textures of the garment. Recently, some methods~\cite{chen2024magic, wang2024stablegarment, shen2024imagdressing} focus on garment-centric image generation. Most of them leverage a full copy of diffusion U-Net as the garment encoder named ReferenceNet to maintain fine-grained garment information. DreamFit~\cite{lin2024dreamfit} proposes a lightweight garment encoder, which utilizes trainable LoRA layers to extract garment features instead of finetuning a full copy of the UNet. 
Nevertheless, these methods are tailored exclusively to single items of clothing and lack support for multiple conditions, thus hindering the ability to freely dress in any combination of various garments. 

In this work, our focus is on a new task \textit{\textbf{Multi-Garment Virtual Dressing}}, personalizing a character wearing any combination of target garments according to the customized text prompt or other controls. The task poses several challenges, including: 1) Garment fidelity: preventing confusion among multiple garments while preserving the intricate textures of each; 2) Text-Image consistency: minimizing the influence of multiple garments on irrelevant regions to ensure the faithfulness of the generated images to the text prompts; 3) Plugin compatibility: enabling seamless integration with community control plugins for LDMs. 


To address the aforementioned issues, we propose \textit{\textbf{AnyDressing}}, a novel approach that customizes characters conditioned on any combination of garments and any personalized text prompts. AnyDressing primarily comprises two primary networks named GarmentsNet and DressingNet.
The GarmentsNet leverages a core Garment-Specific Feature Extractor (GFE) module to extract multi-garment detailed features, which utilizes parallelized self-attention layers within a shared U-Net architecture to individually encode garment textures. And we employ LoRA mechanism within the self-attention layers to further reduce the parameter increase associated with the added garments. The GFE module not only avoids clothing blending but also ensures network efficiency, allowing for easy scalability to any number of garments.
The DressingNet employs a Dressing-Attention (DA) mechanism to seamlessly integrate multi-garment features into the denoising process. To ensure that each garment instance focuses specifically on its corresponding region, we further introduce a novel Instance-Level Garment Localization (IGL) learning strategy in DA. This avoids influencing other irrelevant regions in the synthetic image, thus improving fidelity to arbitrary customized text prompts. 
Additionally, to enhance texture details, we design a Garment-Enhanced Texture Learning (GTL) strategy that strengthens the supervision of attire by imposing constraints from perceptual features and high-frequency information.

Extensive experiments show that AnyDressing has certain advantages in the quantitative and qualitative results compared to state-of-the-art methods. Especially, AnyDressing can serve as a plugin compatible with various fine-tuned LDMs, customized LoRAs~\cite{hu2021lora}, and other extensions such as ControlNet~\cite{zhang2023adding} and IP-Adapter~\cite{ye2023ip}, enhancing the diversity and controllability of synthetic images. In summary, our contributions are as follows:

\begin{itemize}
    \item We propose a novel GarmentsNet to efficiently capture multi-garment textures in parallel by employing a core Garment-Specific Feature Extractor. 
    \item We design a novel DressingNet incorporating a Dressing-Attention mechanism and an Instance-Level Garment Localization Learning strategy to accurately inject multi-garment features into their corresponding regions.
    \item We introduce a Garment-Enhanced Texture Learning strategy to effectively enhance the fine-grained texture details in synthetic images. 
    \item Our framework can seamlessly integrate with any community control plugins for diffusion models. Both quantitative and qualitative experimental results demonstrate the superiority of our AnyDressing.
\end{itemize}


\section{Related Work}
\noindent\textbf{Latent Diffusion Models.}
Latent Diffusion Models (LDMs) \cite{rombach2022high} have become widely used in text-to-image generation tasks. Recent advancements have focused on making generated content more stable and controllable. For instance, ControlNet \cite{zhang2023adding} and T2I Adapter \cite{mou2024t2i} introduced additional conditioning modules injecting control into the denoising U-net via extra branches, such as edges and pose. Additionally, large model fine-tuning methods like LoRA \cite{hu2021lora} have significantly enhanced LDMs’ generative capabilities in specific scenarios. In this work, we can integrate with various fine-tuned LDMs and customized LoRAs to enhance the diversity of generated images. 

\noindent\textbf{\textbf{Subject-Driven Image Generation}.}
Subject-driven generation aims to produce content that aligns with the visual features of a reference image. Methods for this task can be categorized into Tuning-based methods \cite{gal2022image, ruiz2023dreambooth, kumari2023multi, gu2024mix} and Tuning-free methods \cite{li2024blip, li2024photomaker, ye2023ip, ma2024subject, xiao2024fastcomposer, kim2024instantfamily, zhang2024ssr, huang2024resolving, wei2024mm}. Tuning-based methods, such as DreamBooth \cite{ruiz2023dreambooth} and Custom-Diffusion \cite{kumari2023multi} require optimizing specific text tokens to represent target concepts using a limited set of subject images. On the other hand, Tuning-free methods generally encode the reference image into feature embeddings. FastComposer \cite{xiao2024fastcomposer} integrates image features into text embeddings, while IP-Adapter\cite{ye2023ip} and SSR-Encoder\cite{zhang2024ssr} integrate image features into the denoising U-net through a decoupled cross-attention mechanism. However, these methods struggle to preserve the fine-grained texture. 

\noindent\textbf{\textbf{Virtual Try-On}.} Virtual Try-On (VTON) aims to synthesize an image of a specific person wearing a desired garment. Early methods \cite{choi2021viton, wang2018toward, morelli2022dress, lee2022high, issenhuth2020not, xie2023gp, he2022style, li2021toward} utilize generative adversarial networks (GANs) with two-stage strategy, which rely on an explicit warping module and struggle to handle complex backgrounds. Recent studies \cite{morelli2023ladi, gou2023taming, kim2024stableviton, xu2403ootdiffusion, choi2024improving} have used pre-trained LDMs as priors for VTON tasks. LADI-VTON \cite{morelli2023ladi} and DCI-VTON \cite{gou2023taming} explicitly deform the clothes and then use diffusion models to fuse and refine them. Rencent works \cite{kim2024stableviton, xu2403ootdiffusion, choi2024improving} employ parallel U-Nets for clothing feature extraction and inject features into a denoising U-Net. However, VTON is essentially a localized image editing task and requires an existing model image, lacking flexibility in application scenarios.

\noindent\textbf{\textbf{Virtual Dressing}.} Virtual Dressing (VD) \cite{chen2024magic, wang2024stablegarment, shen2024imagdressing} aims to generate freely editable human images with reference garments and optional conditions. StableGarment \cite{wang2024stablegarment} and IMAGDressing \cite{shen2024imagdressing} leverage a garment U-Net for extracting fine-grained clothing features and a denoising U-Net with a hybrid attention module to incorporate garment features into denoising process. Magic Clothing \cite{chen2024magic} additionally proposes a joint classifier-free guidance to balance the control of garment features and text prompts. DreamFit~\cite{lin2024dreamfit} proposes a lightweight garment encoder based on trainable LoRA layers to streamline model complexity and memory usage. 
However, existing approaches are limited to processing single items of clothing, and difficult to maintain fidelity to text prompts. In contrast, our method allows for freely dressing multiple garments and produces coherent and attractive images following customized text prompts.

\section{Preliminaries}
\noindent \textbf{Stable Diffusion.} The Diffusion Model belongs to a class of generative models that generate data through iterative denoising from random noise. In this paper, we specifically employ Stable Diffusion~\cite{rombach2022high}. Stable Diffusion is a latent diffusion model that operates in the latent space of an autoencoder $\mathcal{D}(\mathcal{E}(\cdot))$, where $\mathcal{E}$ and $\mathcal{D}$ represent the encoder and decoder, respectively.
For a given image $\textbf{x}_0$ with its corresponding latent feature $\textbf{z}_0=\mathcal{E}(\textbf{x}_0)$, the diffusion forward process is defined as:
\begin{equation} \label{eauation: noise}
    \textbf{z}_t = \sqrt{\alpha_t} \textbf{z}_0 + \sqrt{1-\alpha_t} \epsilon, 
\end{equation}
where $\alpha_t = \prod_{s=1}^t(1-\beta_s)$, $\epsilon \sim \mathcal{N}(0, 1)$, and $\beta_s$ is the pre-defined variance schedule at timestep $s$. 

In the diffusion backward process, a U-Net $\epsilon_\theta$ is trained to predict the noise. Given the textual condition $\mathbf{C}$, the training objective $\mathcal{L}_{LDM}$ is defined as follows:
\begin{equation} \label{equation: SD}
    \mathcal{L}_{LDM} = \mathbb{E}_{\mathbf{z}_0, \epsilon, \mathbf{C}, t} \Vert \epsilon - \epsilon_\theta(\mathbf{z}_t, \mathbf{C}, t) \Vert_2. 
\end{equation}


\begin{figure*}[ht]
  \centering
  \includegraphics[width=\linewidth]{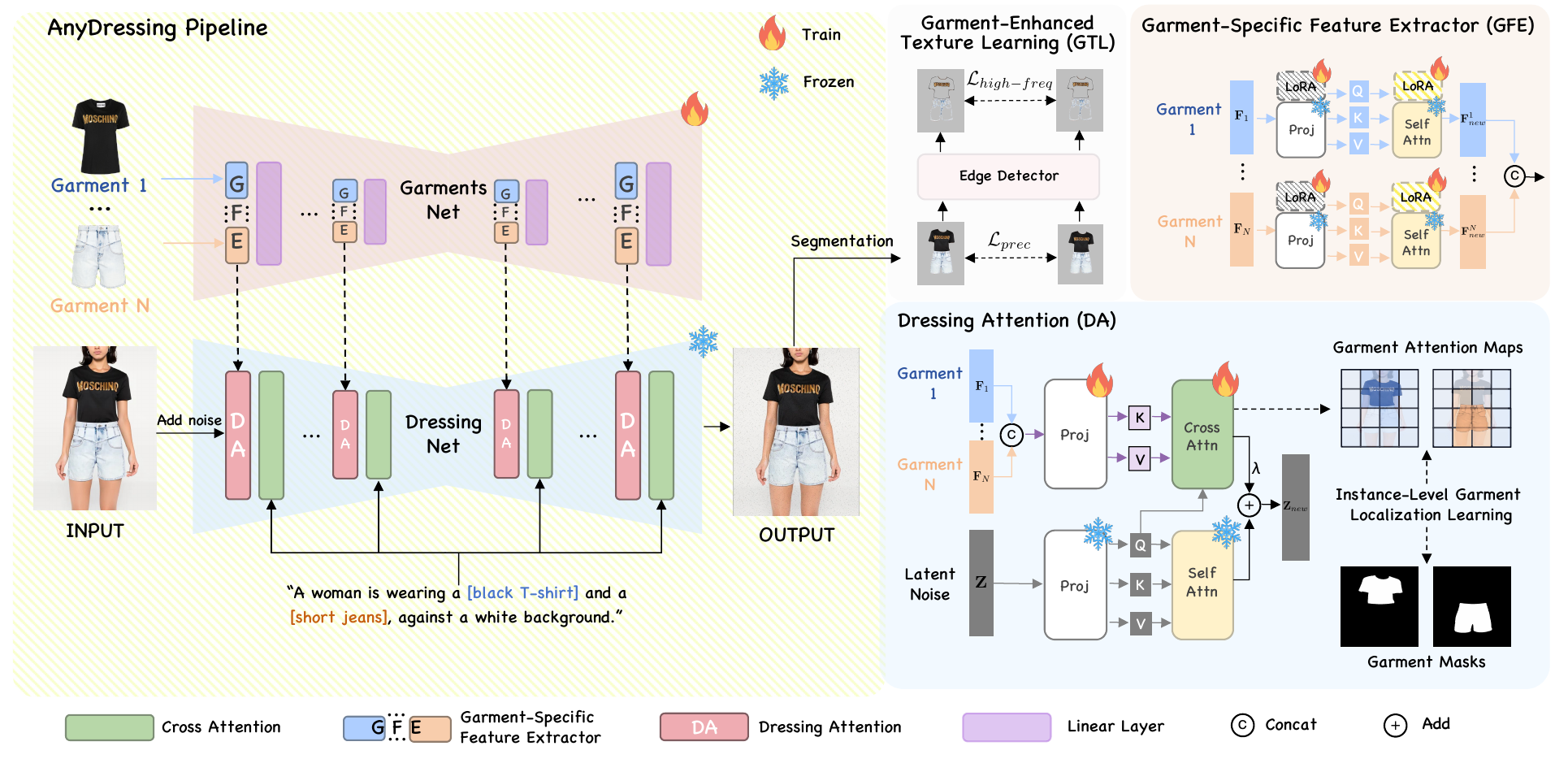}
  \vspace{-0.15in}
  \caption{
    \textbf{Overview of AnyDressing.} Given $N$ target garments, AnyDressing customizes a character dressed in multiple target garments. The GarmentsNet leverages the Garment-Specific Feature Extractor (GFE) module to extract detailed features from multiple garments. The DressingNet integrates these features for virtual dressing using a Dressing-Attention (DA) module and an Instance-Level Garment Localization Learning mechanism. Moreover, the Garment-Enhanced Texture Learning (GTL) strategy further enhances texture details. 
  }
  \vspace{-0.15in}
  \label{fig:architecture}
\end{figure*} 

\section{Methodology}
Given $N$ target garments, the proposed AnyDressing aims to generate a new image $x_{dr}$, showcasing a customized character dressed in multiple target garments across various scenes, styles and actions based on the text prompt. As illustrated in Fig.~\ref{fig:architecture}, AnyDressing comprises two primary networks: GarmentsNet and DressingNet. The GarmentsNet leverages the Garment-Specific Feature Extractor module to extract detailed features from multiple garments (Sec.~\ref{sec:garmentsnet}). Meanwhile, the DressingNet integrates these features for virtual dressing using a Dressing-Attention module and an Instance-Level Garment Localization Learning mechanism (Sec.~\ref{sec:dressingnet}). Additionally, a Garment-Enhanced Texture Learning strategy is designed further to enhance crucial texture details in the synthesis images (Sec.~\ref{sec:detail loss}). Next, we will introduce the aforementioned modules, along with training and inference processes (Sec.~\ref{sec:overall loss}), in detail.



\subsection{GarmentsNet}
\label{sec:garmentsnet}
Previous methods~\cite{chen2024magic, wang2024stablegarment, shen2024imagdressing} leverage a full copy of diffusion U-Net~\cite{chang2023magicdance, hu2024animate} as garment encoding network, ensuring precise preservation of clothing details. However, these methods are limited to handling a single garment and face significant garment confusion issues when applied to multi-garment virtual dressing, as shown in Fig.~\ref{fig:compare}. A straightforward approach to dress multiple garments is to simply duplicate several garment encoding networks to manage different conditions. However, this method would result in a substantial increase in the number of parameters, making it computationally impractical. 

Drawing inspiration from the successful practice of the aforementioned reference mechanisms, we observe that self-attention layers are crucial for the implicit warping of features, enabling the effective matching of input garments to the appropriate body parts. Meanwhile, other layers are typically responsible for general feature extraction and can be shared across different garments without compromising the model’s performance. Building on this insight, we innovatively design a simple yet effective architecture named GarmentsNet, which employs a core Garment-Specific Feature Extractor (GFE) module to encode features for each garment utilizing individual self-attention layers within a shared U-Net framework. 
Inspired by~\cite{lin2024dreamfit}, we integrate LoRA~\cite{hu2021lora} mechanism into self-attention layers, minimizing the increase in parameters associated with the added garments. As a result, this design significantly avoids garment blending while maintaining network efficiency. 
As illustrated in Fig.~\ref{fig:architecture}, the GFE module employs a parallelized self-attention mechanism to extract detailed features of multiple garments. Specifically, for each garment condition $\mathbf{F}_i$, we define the proprietary self-attention LoRA matrix $\triangle \hat{\mathbf{W}}_i$ as follows: 
\begin{equation}
    \triangle \hat{\mathbf{W}}_i = \{ \triangle \hat{\mathbf{W}}_{q}^i, \triangle \hat{\mathbf{W}}_{k}^i, \triangle \hat{\mathbf{W}}_{v}^i \}, 
\end{equation}
where $\triangle \hat{\mathbf{W}}_{q}^i$, $\triangle \hat{\mathbf{W}}_{k}^i$ and $\triangle \hat{\mathbf{W}}_{v}^i$ represent LoRA layers for the query, key and value projections of self-attention layers. We then concatenate self-attention results of each garment condition to obtain the aggregated garment features $\textbf{F}_{new}$:
\begin{equation}
    \mathbf{F}_{new}^i = Softmax(\frac{\mathbf{Q}_i(\mathbf{K}_i)^\top}{\sqrt{d}})\mathbf{V}_i, 
\end{equation}
\begin{equation}
    \mathbf{F}_{new} = Concat (\mathbf{F}_{new}^1, \mathbf{F}_{new}^2, \cdots, \mathbf{F}_{new}^N), 
\end{equation}
where $\mathbf{Q}_i = \mathbf{F}_i (\hat{\mathbf{W}}_{q}+\triangle\hat{\mathbf{W}}_{q}^i)$, $\mathbf{K}_i = \mathbf{F}_i(\hat{\mathbf{W}}_{k}+\triangle\hat{\mathbf{W}}_{k}^i)$, $\mathbf{V}_i = \mathbf{F}_i(\hat{\mathbf{W}}_{v}+\triangle\hat{\mathbf{W}}_{v}^i)$, only $\triangle\hat{\mathbf{W}}$ is trainable and $N$ represents the number of reference garments. 


Thanks to the multi-garment parallel processing design of our GFE module, GarmentsNet can seamlessly scale to any number of garments. Notably, this expansion requires only some newly added LoRA matrix $\triangle \hat{\mathbf{W}}$ in self-attention layers, and significantly minimizes both training and inference time compared with duplicating the entire garment encoding network. Considering the capability of the GFE module to individually encode each garment, we excise the cross-attention modules in GarmentsNet to further reduce redundancy.


\subsection{DressingNet}
\label{sec:dressingnet}
To incorporate multi-garment features during the diffusion process, we meticulously design the DressingNet, which serves as the main denoising net and primarily includes an adaptive Dressing-Attention mechanism and an Instance-Level Garment Localization Learning strategy.

\subsubsection{Adaptive Dressing-Attention}
In the VD task, the main denoising network is typically kept frozen during training~\cite{chen2024magic, shen2024imagdressing} to preserve its original editing and generation capabilities as much as possible. 
To incorporate reference garment features into latent features, we design an adaptive Dressing-Attention (DA) mechanism to efficiently integrate multi-garment texture cues into synthetic images, inspired by~\cite{ye2023ip}. As shown in Fig.~\ref{fig:architecture}, the Dressing-Attention module includes a frozen self-attention module and a learnable cross-attention module. Let $\lbrace \mathbf{F}_1, \mathbf{F}_2, \cdots, \mathbf{F}_N \rbrace$ denote $N$ garment features output by the GarmentsNet at corresponding positions, we first concatenate these features along the spatial dimension to obtain the final garment features: $\mathbf{F}_{all} = Concat ( \mathbf{F}_1, \mathbf{F}_2, \cdots, \mathbf{F}_N)$. 
We then introduce two trainable linear projection layers $\mathbf{W}_{k}^\prime$ and $\mathbf{W}_{v}^\prime$ to align garment features with latent feature $\mathbf{Z}$. Formally, the output of Dressing-Attention $\mathbf{Z}_{new}$ is: 
\begin{equation}
    \mathbf{Z}_{new} = Softmax(\frac{\mathbf{Q}\mathbf{K}^\top}{\sqrt{d}})\mathbf{V} + \lambda * Softmax(\frac{\mathbf{Q}(\mathbf{K^\prime})^\top}{\sqrt{d}})\mathbf{V^\prime}    
\end{equation}
where $\lambda$ is a hyperparameter ensuring the flexibility of incorporating garment features, and $\mathbf{Q} = \mathbf{Z}\mathbf{W}_{q}$, $\mathbf{K} = \mathbf{Z}\mathbf{W}_{k}$, $\mathbf{V} = \mathbf{Z}\mathbf{W}_{v}$, $\mathbf{K}^\prime = \mathbf{F}_{all}\mathbf{W}_{k}^\prime$, $\mathbf{V}^\prime = \mathbf{F}_{all}\mathbf{W}_{v}^\prime$. Here, $\mathbf{W}_{q}$, $\mathbf{W}_{k}$ and $\mathbf{W}_{v}$ are frozen self-attention layers. 
To accelerate the coverage, we initialize the $\mathbf{W}_{k}^\prime$, $\mathbf{W}_{v}^\prime$ with $\mathbf{W}_{k}$, $\mathbf{W}_{v}$. 

\subsubsection{Instance-Level Garment Localization Learning}
\label{subsec:dressing localization}
Although the above Dressing-Attention (DA) mechanism facilitates the integration of multi-garment features, it may result in text-image inconsistency. 
We argue that this results from the garment's attention map covering the entire image in the DA module, thereby injecting garment cues into the other irrelevant regions incorrectly. To tackle this issue, we introduce an Instance-Level Garment Localization (IGL) learning strategy, ensuring that each garment instance focuses solely on its corresponding region. 
Specifically, for each garment feature, we obtain its attention map $A$ with the latent noise in each layer of the DA module: 
\begin{align}
    P = Soft&max(\mathbf{Q}(\mathbf{K}^\prime)^\top / \sqrt{d}), \\
    A &= \sum_{j=1}^{L}P_j ,
\end{align}
where $L$ denotes the length of corresponding garment features. Then, a regularization term $L_{loc}$ is applied to explicitly learn attention localization for each garment instance:
\begin{equation}
    \mathcal{L}_{loc} = \frac{1}{N}\sum_{k=1}^N \Vert A_k - M_k \Vert_2 , 
\end{equation}
where $N$ is the number of garments in the reference image, and $M_k$ represents the reference garment's segmentation mask. 
It is worth noting that the proposed IGL learning strategy is 
applied exclusively during the training phase and does not introduce any additional cost during inference. 

\setlength{\tabcolsep}{1.7mm}{
\begin{table*}
\small
\centering
\renewcommand{\arraystretch}{0.9}
\begin{tabular}{l|ccc|ccc|ccc} 
\Xhline{1pt}
\multirow{3}{*}{Method} & \multicolumn{6}{c|}{\textbf{Single Grament}} & \multicolumn{3}{c}{\textbf{Multiple Graments}} \\
\cline{2-10}
 & \multicolumn{3}{c|}{VITON-HD~\cite{choi2021viton}}  & \multicolumn{3}{c|}{Proprietary Dataset} & \multicolumn{3}{c}{Dressing-Pair}\\
 \cline{2-10}
 & {CLIP-T $\uparrow$} & {CLIP-I} $\uparrow$ & CLIP-AS $\uparrow$ & CLIP-T $\uparrow$ & CLIP-I $\uparrow$ & CLIP-AS $\uparrow$ & CLIP-T $\uparrow$  & CLIP-$\text{I}^*$ $\uparrow$ & CLIP-AS $\uparrow$\\
\midrule
IP-Adapter~\cite{ye2023ip}   & 0.268 & \cellcolor[HTML]{FFFFB2}0.644 & \cellcolor[HTML]{FFFFB2}5.674 & 0.272 & \cellcolor[HTML]{FFFFB2}0.632 & \cellcolor[HTML]{FFFFB2}5.678 & \cellcolor[HTML]{FFFFB2}0.277 & 0.523 & \cellcolor[HTML]{FFD8B2}5.795 \\
StableGarment~\cite{wang2024stablegarment}   & \cellcolor[HTML]{FFFFB2}0.285 & 0.583 & \cellcolor[HTML]{FFD8B2}5.781 & \cellcolor[HTML]{FFFFB2}0.281 & 0.587 & 5.648 & \cellcolor[HTML]{FFD8B2}0.284 & 0.556 & \cellcolor[HTML]{FFFFB2}5.735 \\
MagicClothing~\cite{chen2024magic}   & \cellcolor[HTML]{FFD8B2}0.288 & 0.640 & 5.703 & \cellcolor[HTML]{FFB2B2}0.298 & 0.619 & \cellcolor[HTML]{FFD8B2}5.784 & 0.266 & \cellcolor[HTML]{FFFFB2}0.583 & 5.540 \\
IMAGDressing~\cite{shen2024imagdressing}    & 0.202 & \cellcolor[HTML]{FFD8B2}0.734 & 5.077 & 0.230 & \cellcolor[HTML]{FFD8B2}0.684 & 5.133 & 0.242 & \cellcolor[HTML]{FFD8B2}0.614 & 5.291 \\
$\textbf{Ours}$ & \cellcolor{tabfirst}0.289 & \cellcolor[HTML]{FFB2B2}0.741 & \cellcolor[HTML]{FFB2B2}5.881 & \cellcolor[HTML]{FFD8B2}0.296 & \cellcolor[HTML]{FFB2B2}0.710 & \cellcolor[HTML]{FFB2B2}5.931 & \cellcolor[HTML]{FFB2B2}0.296 & \cellcolor[HTML]{FFB2B2}0.734 & \cellcolor{tabfirst}5.874 \\
\Xhline{1pt}
\end{tabular}
\vspace{-1mm}
\caption{ \textbf{Quantitative comparisons} with baseline methods for both single-garment and multi-garment evaluation. 
}
\label{tab: comp_single}
\end{table*}
}

\subsection{Garment-Enhanced Texture Learning}
\label{sec:detail loss}
Generally, diffusion models are merely optimized relying on the mean-squared loss defined in Eqn.~\ref{equation: SD}, which treats all regions of the synthetic image equally, resulting in a struggle to maintain garment consistency, especially in cases of small text and intricate patterns. To synthesize fine-grained textures, we design a Garment-Enhanced Texture Learning (GTL) strategy to strengthen the supervision of attire details in image space, incorporating a perceptual loss $\mathcal{L}_{perc}$ and a high-frequency loss $\mathcal{L}_{high-freq}$. 

Before introducing the proposed two losses, we define the generated image as: $\hat{I} = \mathcal{D}(\hat{\mathbf{z}}_0)$, where $\mathcal{D}$ denotes the VAE decoder, and $\hat{\mathbf{z}}_0$ is estimated through a single step of inference from the latent $\mathbf{z}_t$:
\begin{equation}
    \hat{\mathbf{z}}_0 = \frac{\mathbf{z}_t - \sqrt{1-\alpha_t}\epsilon_\theta}{\sqrt{\alpha_t}}. 
\end{equation}
Considering the one-step inference may produce noisy and flawed images, the proposed losses are only applied at less noisy timestep ($t \leq \eta$). To sum up, GTL can be defined as:
\begin{equation} 
  \mathcal{L}_{texture} = \begin{cases}
    \mathcal{L}_{perc} + \mathcal{L}_{high-freq}, \ & t \leq \eta \\
        0, \ & t > \eta
  \end{cases}.
\end{equation}


\noindent \textbf{Perception Loss}
To simultaneously enhance structural consistency and pattern similarity with reference garments, we employ a perceptual loss based on the Deep Image Structure and Texture Similarity (DISTS) metric~\cite{ding2020image}. Specifically, we use the reference garment's segmentation mask to isolate the attire in both the generated and ground truth images, averaging their structural and textural inconsistencies within a perceptual feature space, defined as:
\begin{equation}
    \mathcal{L}_{prec} = \frac{1}{N} \sum_{k=1}^N \mathcal{DISTS}(\hat{I} \odot M_k, I \odot M_k), 
\end{equation}
where $\odot$ signifies element-wise multiplication.  

\noindent \textbf{High-Frequency Loss} 
As intricate details in the dressing garments typically appear as high-frequency components with rich edge information, we use edge detection to extract this high-frequency information, aiming to strengthen the constraints on detailed patterns. 
Specifically, we utilize Canny edge detection operator~\cite{ding2001canny} to capture these rich-texture regions, and define the high-frequency loss $\mathcal{L}_{high-freq}$ as:
\begin{equation}
    \mathcal{L}_{high-freq} = \frac{1}{N} \sum_{k=1}^N \Vert \hat{I} \odot M_k^\prime - I \odot M_k^\prime \Vert_2, 
\end{equation}
where $M_k^\prime = M_k \odot P$, $P$ is the extracted edge map of $I$. 

\subsection{Training and Inference}
\label{sec:overall loss}
In training, we average $\mathcal{L}_{loc}$ across all $m$ layers and define overall loss $\mathcal{L}$ as follows:
\begin{align}
    \mathcal{L}_{LDM} &= \mathbb{E}_{\mathbf{z}_0, \epsilon, \mathbf{C}_t, \mathbf{C}_g, t} \Vert \epsilon - \epsilon_\theta(\mathbf{z}_t, \mathbf{C}_t, \mathbf{C}_g, t) \Vert_2, \\
    \mathcal{L} &= \mathcal{L}_{LDM} + \frac{\lambda_1}{m} \mathcal{L}_{loc} + \lambda_2 \mathcal{L}_{texture}, 
\end{align}
where $\mathbf{C}_t$ and $\mathbf{C}_g$ represent text condition and clothing condition respectively. In the inference stage, we apply classifier-free guidance during the denoising process: 
\begin{equation}
    \hat{\epsilon}_\theta(\mathbf{z}_t, \mathbf{C}_t, \mathbf{C}_g, t) = \omega \epsilon_\theta(\mathbf{z}_t, \mathbf{C}_t, \mathbf{C}_g, t) + (1-\omega) \epsilon_\theta(\mathbf{z}_t, t). 
\end{equation}

\section{Experiments}
\subsection{Setup}
\textbf{Dataset.} 
\begin{figure*}[!t]
    \centering
    \includegraphics[width=\textwidth]{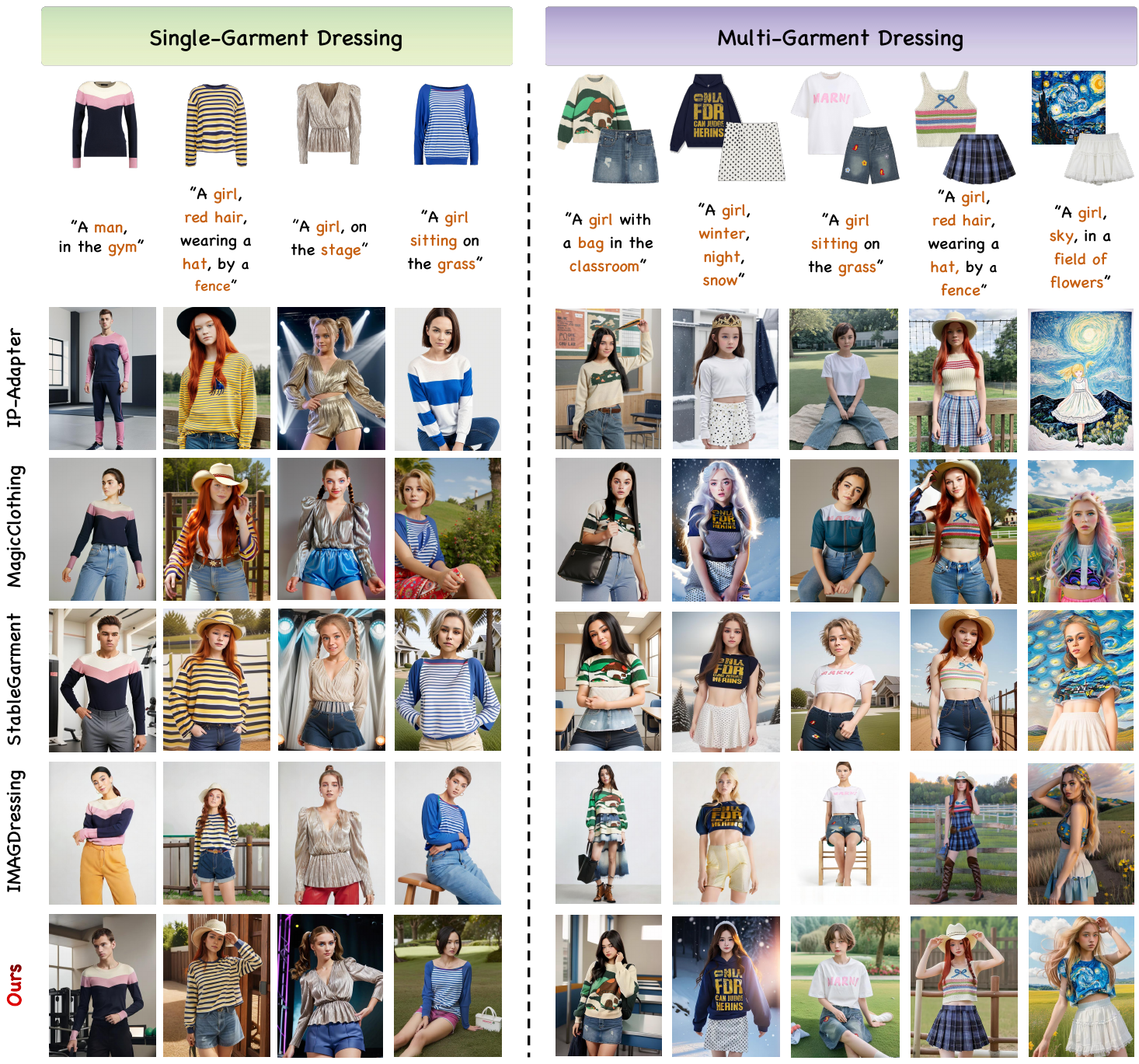}
    \vspace{-0.3cm}
    \caption{\textbf{Qualitative comparisons} with state-of-the-art methods. Please zoom in for more details.}
    \label{fig:compare}
    \vspace{-0.2cm}
\end{figure*}
Notably, a dataset comprising image triplets that include model images paired with multiple laid-out garments is currently lacking. Therefore, we utilize a HumanParsing model to extract clothing items from DressCode~\cite{morelli2022dress} and an additional proprietary dataset collected from the internet, forming triplet data pairs (upper garment, lower garment, person image). In these triplets, one garment is an original laid-out image, while the other is a segmented image from the person’s image. Finally, we construct 26,114 public triplets from Dresscode and 37,065 triplets from proprietary dataset to train AnyDressing. For model evaluation, we introduce two benchmarks to evaluate the model on single-garment and multi-garment dressing respectively. Specifically, for single-garment evaluation, we select 300 reference garments from VITON-HD~\cite{choi2021viton} encompassing various styles and colors, and additionally collect 300 diverse garments with intricate textures from the internet. For multi-garment evaluation, we meticulously gather 25 lowers from the internet and pair each with 10 different uppers, resulting in a total of 250 pairs, called Dressing-Pair. We generate images for each test garment with the provided 7 text prompts. 


\setlength{\tabcolsep}{0.3mm}{
\begin{table}[t]
\small
\centering
\renewcommand{\arraystretch}{0.85}
\resizebox{0.49\textwidth}{!}{
\begin{tabular}{l|cccc}
\Xhline{1pt}
        \small Method & \begin{tabular}[c]{@{}c@{}} \scriptsize Texture \vspace{-2pt} \\ \scriptsize Consistency $\uparrow$ \end{tabular} & \begin{tabular}[c]{@{}c@{}} \scriptsize Align with \vspace{-2pt} \\ \scriptsize Prompt $\uparrow$ \end{tabular} & \begin{tabular}[c]{@{}c@{}} \scriptsize Image \vspace{-2pt} \\ \scriptsize Quality $\uparrow$ \end{tabular} & \begin{tabular}[c]{@{}c@{}} \scriptsize Comprehensive \vspace{-2pt} \\ \scriptsize Evaluation $\uparrow$ \end{tabular} \\ 
\midrule
IP-Adapter~\cite{ye2023ip} & 0.45\% & \cellcolor[HTML]{FFFFB2}6.65\% & \cellcolor[HTML]{FFD8B2}11.95\% & \cellcolor[HTML]{FFFFB2}2.20\% \\
StableGarment~\cite{wang2024stablegarment} & 1.60\% & 4.85\% & 2.65\% & 2.05\% \\
MagicClothing~\cite{chen2024magic} & \cellcolor[HTML]{FFFFB2}2.05\% & \cellcolor[HTML]{FFD8B2}9.00\% & \cellcolor[HTML]{FFFFB2}9.70\% & \cellcolor[HTML]{FFD8B2}3.75\% \\
IMAGDressing~\cite{shen2024imagdressing} & \cellcolor[HTML]{FFD8B2}2.10\% & 2.50\% & 3.90\% & 1.70\% \\
\textbf{Ours} & \cellcolor[HTML]{FFB2B2}93.80\% & \cellcolor[HTML]{FFB2B2}77.00\% & \cellcolor[HTML]{FFB2B2}71.80\% & \cellcolor[HTML]{FFB2B2}90.30\% \\ 
\Xhline{1pt}
\end{tabular}
}
\vspace{-1mm}
\caption{\small \textbf{User study} with baseline methods.}
\vspace{-5mm}
\label{tab: user_study}
\end{table}
}

\noindent\textbf{Implementation Details.} In our experiments, we initialize the weights of GarmentsNet and DressingNet with the weights of the U-Net in Stable Diffusion v1.5~\cite{rombach2022high}. Our model is trained on paired images at the resolution of $768 \times 576$. The trainable parameters are GarmentsNet and the cross-attention layers in Dressing-Attention module. During training, We adopt AdamW~\cite{loshchilov2017decoupled} optimizer with a fixed learning rate of 5e-5. The model is trained for 100k steps on 8 NVIDIA A100 GPUs with a batch size of 4. During inference, we use DDIM~\cite{song2020denoising} sampler with 30 steps and set guidance scale $\omega$ to 6.0. Please refer to the supplementary materials for more details.

\noindent\textbf{Baselines.} 
We compare our method against the following state-of-the-art image synthesis method: IP-Adapter~\cite{ye2023ip}, MagicClothing~\cite{chen2024magic}, StableGarment~\cite{wang2024stablegarment} and IMAGDressing~\cite{shen2024imagdressing}. We use the official model parameters from their official implementations. For a fair comparison, all experiments are conducted with the resolution of $768 \times 576$. 

\noindent\textbf{Evaluation Metrics.} We follow previous methods to adopt
three metrics for evaluation: CLIP-T for text-image similarity, CLIP-I for garment consistency, and CLIP Aesthetic Score (CLIP-AS) for overall generation quality. Especially, to better evaluate multi-garment dressing, we introduce a new metric CLIP-$\text{I}^*$ to assess texture consistency by leveraging OpenPose~\cite{cao2020openpose} to obtain the matching partitions of the reference garments in the synthesized image and averaging their CLIP-I metrics. 

\subsection{Qualitative Analysis}
\begin{figure}[!t]
    \centering
    \includegraphics[width=0.47\textwidth]{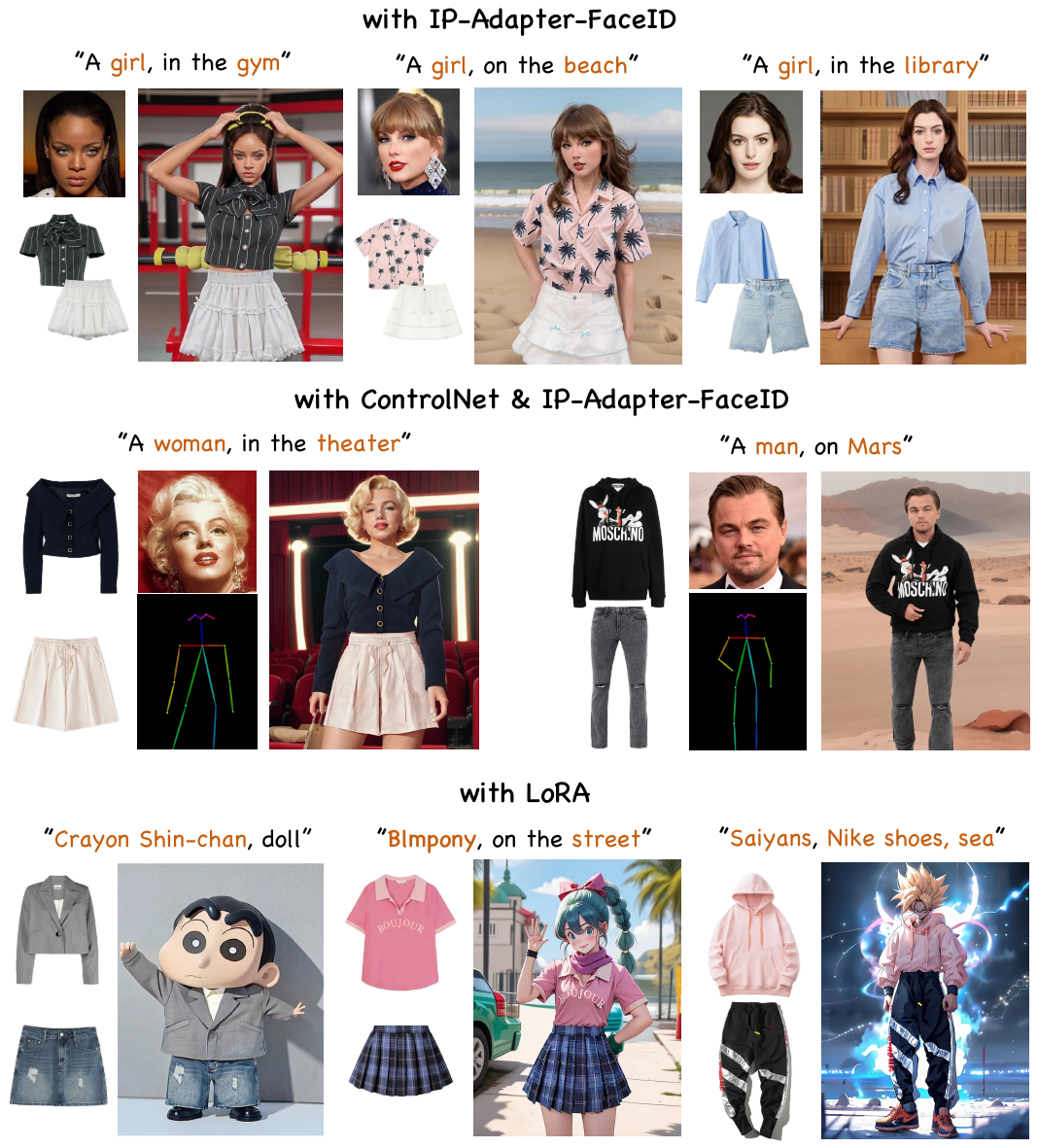}
    \caption{ \small \textbf{Examples of plug-in} results of AnyDressing.}
    \label{fig:plug-in}
    \vspace*{-1mm}
\end{figure}

Since the compared methods lack multi-garment support, we obtain baseline results by concatenating multiple garments along the spatial dimension as input. Fig.~\ref{fig:compare} presents visual comparisons between our method and baseline approaches. AnyDressing maintains superior consistency in clothing style and texture, and exhibits better text fidelity, while other methods struggle to balance garment preservation and prompt faithfulness. In particular, baselines encounter significant background contamination and garment confusion in multi-garment dressing results, whereas our method demonstrates exceptional reliability, which is attributed to our designed GarmentsNet and DressingNet architectures. 
And Fig.~\ref{fig:plug-in} presents the results of AnyDressing as a plug-in module combined with other extensions and customized LoRAs, demonstrating its powerful compatibility. Please refer to the supplementary for more results. 

\subsection{Quantitative Comparisons}
\textbf{Metric Evaluation.} Tab.~\ref{tab: comp_single} shows the quantitative results of our methods against baselines. For single-garment evaluation, extensive experiments conducted on VITON-HD~\cite{choi2021viton} and proprietary dataset prove the superiority of AnyDressing compared with all baselines. And our method significantly surpasses all baselines across all metrics in multi-garment virtual dressing results, fully demonstrating AnyDressing's reliability in handling both single-garment and multi-garment virtual dressing tasks.

\noindent \textbf{User Study.} We conduct a user study to evaluate the generation quality of our model. We use all test garments and prompts in our dataset and randomly show the users $25$ single-garment results and $25$ multi-garment results from the baselines and our method. Each participant is asked to select the \textbf{\textit{most}} preferred result under four criteria: texture consistency, alignment with the text prompt, image quality and comprehensive evaluation. In the end, we receive valid responses from $40$ users. The collected preferences are reported in Tab.~\ref{tab: user_study}. In terms of four criteria, our method is preferred by most participants, with percentages reaching $93.80\%$, $77.00\%$, $71.80\%$ and $90.30\%$ respectively. 

\begin{figure}[!t]
    \centering
    \includegraphics[width=0.47\textwidth]{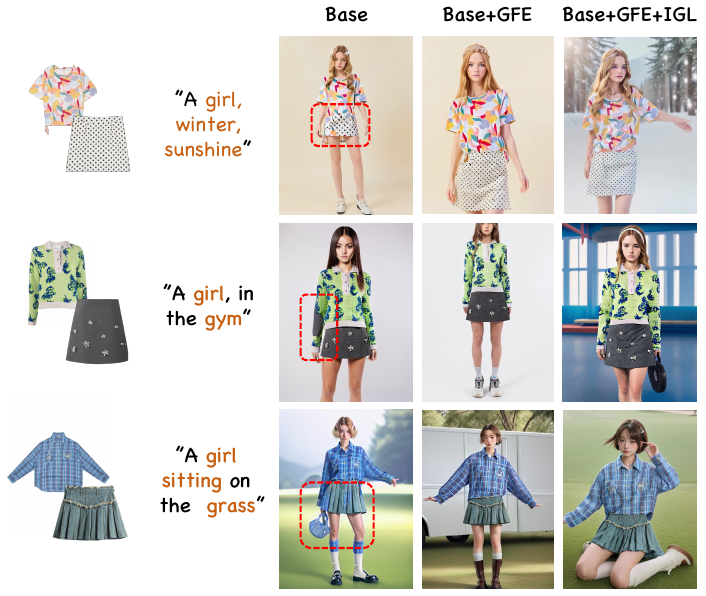}
    \caption{\small \textbf{Ablation results} on GFE and IGL modules.}
    \label{fig:abla_1}
    \vspace*{-3mm}
\end{figure}
\begin{figure}[!t]
    \centering
    \includegraphics[width=0.47\textwidth]{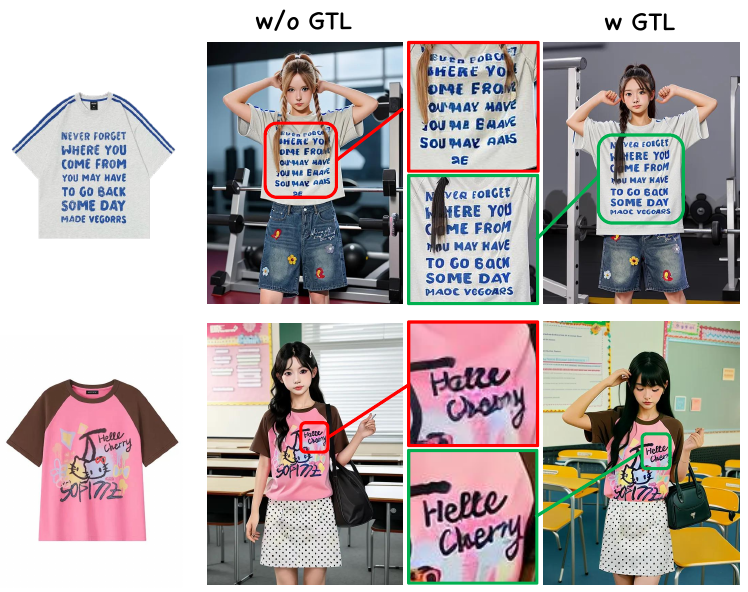}
    \caption{\small \textbf{Ablation results} on GTL module.}
    \label{fig:abla_2}
    \vspace*{-4.5mm}
\end{figure}

\subsection{Ablation Studies}
\label{sec: abl}
\setlength{\tabcolsep}{2.2mm}{
\begin{table}
\small
\centering
\renewcommand{\arraystretch}{0.8}
\begin{tabular}{c c c | c c c} 
\Xhline{1pt}
\small GFE & IGL & GTL & CLIP-T $\uparrow$ & CLIP-$\text{I}^*$ $\uparrow$ & CLIP-AS $\uparrow$\\
\midrule
\textcolor{violet}{\XSolidBrush} & \textcolor{violet}{\XSolidBrush} & \textcolor{violet}{\XSolidBrush}  & 0.260 & 0.625 & 5.572 \\
\textcolor{olive}{\CheckmarkBold}& \textcolor{violet}{\XSolidBrush}& \textcolor{violet}{\XSolidBrush} & 0.265 & 0.718 & 5.627 \\
\textcolor{olive}{\CheckmarkBold}& \textcolor{olive}{\CheckmarkBold}& \textcolor{violet}{\XSolidBrush} & 0.289 & 0.722 & 5.790 \\
\textcolor{olive}{\CheckmarkBold}& \textcolor{olive}{\CheckmarkBold}& \textcolor{olive}{\CheckmarkBold} & \textbf{0.296} & \textbf{0.734} & \textbf{5.874}\\
\Xhline{1pt}
\end{tabular}
\vspace{-1mm}
\caption{ \small \textbf{Ablation study} of AnyDressing. 
}
\vspace{-4.5mm}
\label{tab: abl}
\end{table}
}
\textbf{GFE \& IGL.} To validate the effectiveness of our proposed architecture, we employ traditional ReferenceNet~\cite{hu2024animate} to encode multiple garments concurrently and then incorporate them into the denoising U-Net similar to~\cite{shen2024imagdressing} as our base model. As illustrated in Fig.~\ref{fig:abla_1}, Base+GFE significantly reduces garment confusion and improves garment consistency compared to Base, which is attributed to the multi-garment parallel processing design of the GFE module. Base+GFE+IGL shows better fidelity to the text prompts and further mitigates background contamination, which demonstrates IGL mechanism effectively constrains garment features to attend to the correct regions. The quantitative comparison in Tab.~\ref{tab: abl} further proves the effectiveness of each module, with GFE primarily improving the CLIP-$\text{I}^*$ and IGL enhancing both CLIP-T and CLIP-AS.

\noindent\textbf{GTL.} Fig.~\ref{fig:abla_2} intuitively demonstrates the effectiveness of our proposed GTL strategy, encouraging the model to enhance detail preservation, particularly in small text and intricate patterns. And quantitative result in Tab.~\ref{tab: abl} also verifies that our designed GTL improves texture consistency. 

\section{Conclusion}
This paper presents AnyDressing comprising two core networks named GarmentsNet and DressingNet to focus on a new task, i.e., Multi-Garment Virtual Dressing. The GarmentsNet employs a Garment-Specific Feature Extractor module to efficiently encode multi-garment features in parallel. The DressingNet integrates these features for virtual dressing using a Dressing-Attention module and an Instance-Level Garment Localization Learning mechanism. Additionally, we design a Garment-Enhanced Texture Learning strategy to further enhance texture details. Our approach can seamlessly integrate with any community control plugins. Extensive experiments show that AnyDressing achieves state-of-the-art results. 
{
    \small
    \bibliographystyle{ieeenat_fullname}
    \bibliography{main}

\begin{thebibliography}{57}
\providecommand{\natexlab}[1]{#1}
\providecommand{\url}[1]{\texttt{#1}}
\expandafter\ifx\csname urlstyle\endcsname\relax
  \providecommand{\doi}[1]{doi: #1}\else
  \providecommand{\doi}{doi: \begingroup \urlstyle{rm}\Url}\fi

\bibitem[Cao et~al.(2020)Cao, Hidalgo, Simon, Wei, and Sheikh]{cao2020openpose}
Z Cao, G Hidalgo, T Simon, SE Wei, and Y Sheikh.
\newblock Openpose: Realtime multi-person 2d pose estimation using part affinity fields.
\newblock \emph{IEEE Transactions on Pattern Analysis and Machine Intelligence}, 43\penalty0 (1):\penalty0 172--186, 2020.

\bibitem[Chang et~al.(2023{\natexlab{a}})Chang, Shi, Gao, Fu, Xu, Song, Yan, Yang, and Soleymani]{chang2023magicdance}
Di Chang, Yichun Shi, Quankai Gao, Jessica Fu, Hongyi Xu, Guoxian Song, Qing Yan, Xiao Yang, and Mohammad Soleymani.
\newblock Magicdance: Realistic human dance video generation with motions \& facial expressions transfer.
\newblock \emph{arXiv preprint arXiv:2311.12052}, 2023{\natexlab{a}}.

\bibitem[Chang et~al.(2023{\natexlab{b}})Chang, Zhang, Barber, Maschinot, Lezama, Jiang, Yang, Murphy, Freeman, Rubinstein, et~al.]{chang2023muse}
Huiwen Chang, Han Zhang, Jarred Barber, AJ Maschinot, Jose Lezama, Lu Jiang, Ming-Hsuan Yang, Kevin Murphy, William~T Freeman, Michael Rubinstein, et~al.
\newblock Muse: Text-to-image generation via masked generative transformers.
\newblock \emph{arXiv preprint arXiv:2301.00704}, 2023{\natexlab{b}}.

\bibitem[Chen et~al.(2024)Chen, Gu, Xu, and Chen]{chen2024magic}
Weifeng Chen, Tao Gu, Yuhao Xu, and Chengcai Chen.
\newblock Magic clothing: Controllable garment-driven image synthesis.
\newblock \emph{arXiv preprint arXiv:2404.09512}, 2024.

\bibitem[Choi et~al.(2021)Choi, Park, Lee, and Choo]{choi2021viton}
Seunghwan Choi, Sunghyun Park, Minsoo Lee, and Jaegul Choo.
\newblock Viton-hd: High-resolution virtual try-on via misalignment-aware normalization.
\newblock In \emph{Proceedings of the IEEE/CVF conference on computer vision and pattern recognition}, pages 14131--14140, 2021.

\bibitem[Choi et~al.(2024)Choi, Kwak, Lee, Choi, and Shin]{choi2024improving}
Yisol Choi, Sangkyung Kwak, Kyungmin Lee, Hyungwon Choi, and Jinwoo Shin.
\newblock Improving diffusion models for virtual try-on.
\newblock \emph{arXiv preprint arXiv:2403.05139}, 2024.

\bibitem[Ding et~al.(2020)Ding, Ma, Wang, and Simoncelli]{ding2020image}
Keyan Ding, Kede Ma, Shiqi Wang, and Eero~P Simoncelli.
\newblock Image quality assessment: Unifying structure and texture similarity.
\newblock \emph{IEEE transactions on pattern analysis and machine intelligence}, 44\penalty0 (5):\penalty0 2567--2581, 2020.

\bibitem[Ding and Goshtasby(2001)]{ding2001canny}
Lijun Ding and Ardeshir Goshtasby.
\newblock On the canny edge detector.
\newblock \emph{Pattern recognition}, 34\penalty0 (3):\penalty0 721--725, 2001.

\bibitem[Ding et~al.(2021)Ding, Yang, Hong, Zheng, Zhou, Yin, Lin, Zou, Shao, Yang, et~al.]{ding2021cogview}
Ming Ding, Zhuoyi Yang, Wenyi Hong, Wendi Zheng, Chang Zhou, Da Yin, Junyang Lin, Xu Zou, Zhou Shao, Hongxia Yang, et~al.
\newblock Cogview: Mastering text-to-image generation via transformers.
\newblock \emph{Advances in neural information processing systems}, 34:\penalty0 19822--19835, 2021.

\bibitem[Gal et~al.(2022)Gal, Alaluf, Atzmon, Patashnik, Bermano, Chechik, and Cohen-Or]{gal2022image}
Rinon Gal, Yuval Alaluf, Yuval Atzmon, Or Patashnik, Amit~H Bermano, Gal Chechik, and Daniel Cohen-Or.
\newblock An image is worth one word: Personalizing text-to-image generation using textual inversion.
\newblock \emph{arXiv preprint arXiv:2208.01618}, 2022.

\bibitem[Gou et~al.(2023)Gou, Sun, Zhang, Si, Qian, and Zhang]{gou2023taming}
Junhong Gou, Siyu Sun, Jianfu Zhang, Jianlou Si, Chen Qian, and Liqing Zhang.
\newblock Taming the power of diffusion models for high-quality virtual try-on with appearance flow.
\newblock In \emph{Proceedings of the 31st ACM International Conference on Multimedia}, pages 7599--7607, 2023.

\bibitem[Gu et~al.(2024)Gu, Wang, Wu, Shi, Chen, Fan, Xiao, Zhao, Chang, Wu, et~al.]{gu2024mix}
Yuchao Gu, Xintao Wang, Jay~Zhangjie Wu, Yujun Shi, Yunpeng Chen, Zihan Fan, Wuyou Xiao, Rui Zhao, Shuning Chang, Weijia Wu, et~al.
\newblock Mix-of-show: Decentralized low-rank adaptation for multi-concept customization of diffusion models.
\newblock \emph{Advances in Neural Information Processing Systems}, 36, 2024.

\bibitem[He et~al.(2022)He, Song, and Xiang]{he2022style}
Sen He, Yi-Zhe Song, and Tao Xiang.
\newblock Style-based global appearance flow for virtual try-on.
\newblock In \emph{Proceedings of the IEEE/CVF Conference on Computer Vision and Pattern Recognition}, pages 3470--3479, 2022.

\bibitem[Ho et~al.(2020)Ho, Jain, and Abbeel]{ho2020denoising}
Jonathan Ho, Ajay Jain, and Pieter Abbeel.
\newblock Denoising diffusion probabilistic models.
\newblock \emph{Advances in neural information processing systems}, 33:\penalty0 6840--6851, 2020.

\bibitem[Hu et~al.(2021)Hu, Shen, Wallis, Allen-Zhu, Li, Wang, Wang, and Chen]{hu2021lora}
Edward~J Hu, Yelong Shen, Phillip Wallis, Zeyuan Allen-Zhu, Yuanzhi Li, Shean Wang, Lu Wang, and Weizhu Chen.
\newblock Lora: Low-rank adaptation of large language models.
\newblock \emph{arXiv preprint arXiv:2106.09685}, 2021.

\bibitem[Hu(2024)]{hu2024animate}
Li Hu.
\newblock Animate anyone: Consistent and controllable image-to-video synthesis for character animation.
\newblock In \emph{Proceedings of the IEEE/CVF Conference on Computer Vision and Pattern Recognition}, pages 8153--8163, 2024.

\bibitem[Huang et~al.(2024)Huang, Fu, Liu, Jiang, Yu, and Song]{huang2024resolving}
Qihan Huang, Siming Fu, Jinlong Liu, Hao Jiang, Yipeng Yu, and Jie Song.
\newblock Resolving multi-condition confusion for finetuning-free personalized image generation.
\newblock \emph{arXiv preprint arXiv:2409.17920}, 2024.

\bibitem[Huang et~al.(2023)Huang, Wu, Jiang, Chan, and Liu]{huang2023reversion}
Ziqi Huang, Tianxing Wu, Yuming Jiang, Kelvin~CK Chan, and Ziwei Liu.
\newblock Reversion: Diffusion-based relation inversion from images.
\newblock \emph{arXiv preprint arXiv:2303.13495}, 2023.

\bibitem[Issenhuth et~al.(2020)Issenhuth, Mary, and Calauzenes]{issenhuth2020not}
Thibaut Issenhuth, J{\'e}r{\'e}mie Mary, and Cl{\'e}ment Calauzenes.
\newblock Do not mask what you do not need to mask: a parser-free virtual try-on.
\newblock In \emph{Computer Vision--ECCV 2020: 16th European Conference, Glasgow, UK, August 23--28, 2020, Proceedings, Part XX 16}, pages 619--635. Springer, 2020.

\bibitem[Jin(2023)]{jin2023sssegmenation}
Zhenchao Jin.
\newblock Sssegmenation: An open source supervised semantic segmentation toolbox based on pytorch.
\newblock \emph{arXiv preprint arXiv:2305.17091}, 2023.

\bibitem[Jin et~al.(2024)Jin, Hu, Zhu, Song, Yuan, and Yu]{jin2024idrnet}
Zhenchao Jin, Xiaowei Hu, Lingting Zhu, Luchuan Song, Li Yuan, and Lequan Yu.
\newblock Idrnet: Intervention-driven relation network for semantic segmentation.
\newblock \emph{Advances in Neural Information Processing Systems}, 36, 2024.

\bibitem[Kang et~al.(2023)Kang, Zhu, Zhang, Park, Shechtman, Paris, and Park]{kang2023scaling}
Minguk Kang, Jun-Yan Zhu, Richard Zhang, Jaesik Park, Eli Shechtman, Sylvain Paris, and Taesung Park.
\newblock Scaling up gans for text-to-image synthesis.
\newblock In \emph{Proceedings of the IEEE/CVF Conference on Computer Vision and Pattern Recognition}, pages 10124--10134, 2023.

\bibitem[Kim et~al.(2024{\natexlab{a}})Kim, Lee, Joung, Kim, and Baek]{kim2024instantfamily}
Chanran Kim, Jeongin Lee, Shichang Joung, Bongmo Kim, and Yeul-Min Baek.
\newblock Instantfamily: Masked attention for zero-shot multi-id image generation.
\newblock \emph{arXiv preprint arXiv:2404.19427}, 2024{\natexlab{a}}.

\bibitem[Kim et~al.(2024{\natexlab{b}})Kim, Gu, Park, Park, and Choo]{kim2024stableviton}
Jeongho Kim, Guojung Gu, Minho Park, Sunghyun Park, and Jaegul Choo.
\newblock Stableviton: Learning semantic correspondence with latent diffusion model for virtual try-on.
\newblock In \emph{Proceedings of the IEEE/CVF Conference on Computer Vision and Pattern Recognition}, pages 8176--8185, 2024{\natexlab{b}}.

\bibitem[Kumari et~al.(2023)Kumari, Zhang, Zhang, Shechtman, and Zhu]{kumari2023multi}
Nupur Kumari, Bingliang Zhang, Richard Zhang, Eli Shechtman, and Jun-Yan Zhu.
\newblock Multi-concept customization of text-to-image diffusion.
\newblock In \emph{Proceedings of the IEEE/CVF Conference on Computer Vision and Pattern Recognition}, pages 1931--1941, 2023.

\bibitem[Lee et~al.(2022)Lee, Gu, Park, Choi, and Choo]{lee2022high}
Sangyun Lee, Gyojung Gu, Sunghyun Park, Seunghwan Choi, and Jaegul Choo.
\newblock High-resolution virtual try-on with misalignment and occlusion-handled conditions.
\newblock In \emph{European Conference on Computer Vision}, pages 204--219. Springer, 2022.

\bibitem[Li et~al.(2024{\natexlab{a}})Li, Li, and Hoi]{li2024blip}
Dongxu Li, Junnan Li, and Steven Hoi.
\newblock Blip-diffusion: Pre-trained subject representation for controllable text-to-image generation and editing.
\newblock \emph{Advances in Neural Information Processing Systems}, 36, 2024{\natexlab{a}}.

\bibitem[Li et~al.(2021)Li, Chong, Zhang, and Liu]{li2021toward}
Kedan Li, Min~Jin Chong, Jeffrey Zhang, and Jingen Liu.
\newblock Toward accurate and realistic outfits visualization with attention to details.
\newblock In \emph{Proceedings of the IEEE/CVF conference on computer vision and pattern recognition}, pages 15546--15555, 2021.

\bibitem[Li et~al.(2024{\natexlab{b}})Li, Cao, Wang, Qi, Cheng, and Shan]{li2024photomaker}
Zhen Li, Mingdeng Cao, Xintao Wang, Zhongang Qi, Ming-Ming Cheng, and Ying Shan.
\newblock Photomaker: Customizing realistic human photos via stacked id embedding.
\newblock In \emph{Proceedings of the IEEE/CVF Conference on Computer Vision and Pattern Recognition}, pages 8640--8650, 2024{\natexlab{b}}.

\bibitem[Lin et~al.(2024)Lin, Zhang, Zhao, Luo, Dong, Zeng, and Liang]{lin2024dreamfit}
Ente Lin, Xujie Zhang, Fuwei Zhao, Yuxuan Luo, Xin Dong, Long Zeng, and Xiaodan Liang.
\newblock Dreamfit: Garment-centric human generation via a lightweight anything-dressing encoder.
\newblock \emph{arXiv preprint arXiv:2412.17644}, 2024.

\bibitem[Liu et~al.(2023)Liu, Zhang, Shen, Zheng, Zhu, Feng, Liu, Zhao, Zhou, and Cao]{liu2023cones}
Zhiheng Liu, Yifei Zhang, Yujun Shen, Kecheng Zheng, Kai Zhu, Ruili Feng, Yu Liu, Deli Zhao, Jingren Zhou, and Yang Cao.
\newblock Cones 2: Customizable image synthesis with multiple subjects.
\newblock In \emph{Proceedings of the 37th International Conference on Neural Information Processing Systems}, pages 57500--57519, 2023.

\bibitem[Loshchilov(2017)]{loshchilov2017decoupled}
I Loshchilov.
\newblock Decoupled weight decay regularization.
\newblock \emph{arXiv preprint arXiv:1711.05101}, 2017.

\bibitem[Ma et~al.(2024)Ma, Liang, Chen, and Lu]{ma2024subject}
Jian Ma, Junhao Liang, Chen Chen, and Haonan Lu.
\newblock Subject-diffusion: Open domain personalized text-to-image generation without test-time fine-tuning.
\newblock In \emph{ACM SIGGRAPH 2024 Conference Papers}, pages 1--12, 2024.

\bibitem[Morelli et~al.(2022)Morelli, Fincato, Cornia, Landi, Cesari, and Cucchiara]{morelli2022dress}
Davide Morelli, Matteo Fincato, Marcella Cornia, Federico Landi, Fabio Cesari, and Rita Cucchiara.
\newblock Dress code: High-resolution multi-category virtual try-on.
\newblock In \emph{Proceedings of the IEEE/CVF conference on computer vision and pattern recognition}, pages 2231--2235, 2022.

\bibitem[Morelli et~al.(2023)Morelli, Baldrati, Cartella, Cornia, Bertini, and Cucchiara]{morelli2023ladi}
Davide Morelli, Alberto Baldrati, Giuseppe Cartella, Marcella Cornia, Marco Bertini, and Rita Cucchiara.
\newblock Ladi-vton: Latent diffusion textual-inversion enhanced virtual try-on.
\newblock In \emph{Proceedings of the 31st ACM International Conference on Multimedia}, pages 8580--8589, 2023.

\bibitem[Mou et~al.(2024)Mou, Wang, Xie, Wu, Zhang, Qi, and Shan]{mou2024t2i}
Chong Mou, Xintao Wang, Liangbin Xie, Yanze Wu, Jian Zhang, Zhongang Qi, and Ying Shan.
\newblock T2i-adapter: Learning adapters to dig out more controllable ability for text-to-image diffusion models.
\newblock In \emph{Proceedings of the AAAI Conference on Artificial Intelligence}, pages 4296--4304, 2024.

\bibitem[Ramesh et~al.(2022)Ramesh, Dhariwal, Nichol, Chu, and Chen]{ramesh2022hierarchical}
Aditya Ramesh, Prafulla Dhariwal, Alex Nichol, Casey Chu, and Mark Chen.
\newblock Hierarchical text-conditional image generation with clip latents.
\newblock \emph{arXiv preprint arXiv:2204.06125}, 1\penalty0 (2):\penalty0 3, 2022.

\bibitem[Rombach et~al.(2022)Rombach, Blattmann, Lorenz, Esser, and Ommer]{rombach2022high}
Robin Rombach, Andreas Blattmann, Dominik Lorenz, Patrick Esser, and Bj{\"o}rn Ommer.
\newblock High-resolution image synthesis with latent diffusion models.
\newblock In \emph{Proceedings of the IEEE/CVF conference on computer vision and pattern recognition}, pages 10684--10695, 2022.

\bibitem[Ruiz et~al.(2023)Ruiz, Li, Jampani, Pritch, Rubinstein, and Aberman]{ruiz2023dreambooth}
Nataniel Ruiz, Yuanzhen Li, Varun Jampani, Yael Pritch, Michael Rubinstein, and Kfir Aberman.
\newblock Dreambooth: Fine tuning text-to-image diffusion models for subject-driven generation.
\newblock In \emph{Proceedings of the IEEE/CVF conference on computer vision and pattern recognition}, pages 22500--22510, 2023.

\bibitem[Shen et~al.(2024)Shen, Jiang, He, Ye, Wang, Du, Li, and Tang]{shen2024imagdressing}
Fei Shen, Xin Jiang, Xin He, Hu Ye, Cong Wang, Xiaoyu Du, Zechao Li, and Jinghui Tang.
\newblock Imagdressing-v1: Customizable virtual dressing.
\newblock \emph{arXiv preprint arXiv:2407.12705}, 2024.

\bibitem[Shi et~al.(2024)Shi, Xiong, Lin, and Jung]{shi2024instantbooth}
Jing Shi, Wei Xiong, Zhe Lin, and Hyun~Joon Jung.
\newblock Instantbooth: Personalized text-to-image generation without test-time finetuning.
\newblock In \emph{Proceedings of the IEEE/CVF Conference on Computer Vision and Pattern Recognition}, pages 8543--8552, 2024.

\bibitem[Sohl-Dickstein et~al.(2015)Sohl-Dickstein, Weiss, Maheswaranathan, and Ganguli]{sohl2015deep}
Jascha Sohl-Dickstein, Eric Weiss, Niru Maheswaranathan, and Surya Ganguli.
\newblock Deep unsupervised learning using nonequilibrium thermodynamics.
\newblock In \emph{International conference on machine learning}, pages 2256--2265. PMLR, 2015.

\bibitem[Song et~al.(2020{\natexlab{a}})Song, Meng, and Ermon]{song2020denoising}
Jiaming Song, Chenlin Meng, and Stefano Ermon.
\newblock Denoising diffusion implicit models.
\newblock \emph{arXiv preprint arXiv:2010.02502}, 2020{\natexlab{a}}.

\bibitem[Song et~al.(2020{\natexlab{b}})Song, Sohl-Dickstein, Kingma, Kumar, Ermon, and Poole]{song2020score}
Yang Song, Jascha Sohl-Dickstein, Diederik~P Kingma, Abhishek Kumar, Stefano Ermon, and Ben Poole.
\newblock Score-based generative modeling through stochastic differential equations.
\newblock \emph{arXiv preprint arXiv:2011.13456}, 2020{\natexlab{b}}.

\bibitem[Vinker et~al.(2023)Vinker, Voynov, Cohen-Or, and Shamir]{vinker2023concept}
Yael Vinker, Andrey Voynov, Daniel Cohen-Or, and Ariel Shamir.
\newblock Concept decomposition for visual exploration and inspiration.
\newblock \emph{ACM Transactions on Graphics (TOG)}, 42\penalty0 (6):\penalty0 1--13, 2023.

\bibitem[Wang et~al.(2018)Wang, Zheng, Liang, Chen, Lin, and Yang]{wang2018toward}
Bochao Wang, Huabin Zheng, Xiaodan Liang, Yimin Chen, Liang Lin, and Meng Yang.
\newblock Toward characteristic-preserving image-based virtual try-on network.
\newblock In \emph{Proceedings of the European conference on computer vision (ECCV)}, pages 589--604, 2018.

\bibitem[Wang et~al.(2024)Wang, Guo, Liu, Li, Zhao, Tang, Hu, Tang, and Li]{wang2024stablegarment}
Rui Wang, Hailong Guo, Jiaming Liu, Huaxia Li, Haibo Zhao, Xu Tang, Yao Hu, Hao Tang, and Peipei Li.
\newblock Stablegarment: Garment-centric generation via stable diffusion.
\newblock \emph{arXiv preprint arXiv:2403.10783}, 2024.

\bibitem[Wang et~al.(2023)Wang, Lv, Yu, Hong, Qi, Wang, Ji, Yang, Zhao, Song, et~al.]{wang2023cogvlm}
Weihan Wang, Qingsong Lv, Wenmeng Yu, Wenyi Hong, Ji Qi, Yan Wang, Junhui Ji, Zhuoyi Yang, Lei Zhao, Xixuan Song, et~al.
\newblock Cogvlm: Visual expert for pretrained language models.
\newblock \emph{arXiv preprint arXiv:2311.03079}, 2023.

\bibitem[Wei et~al.(2023)Wei, Zhang, Ji, Bai, Zhang, and Zuo]{wei2023elite}
Yuxiang Wei, Yabo Zhang, Zhilong Ji, Jinfeng Bai, Lei Zhang, and Wangmeng Zuo.
\newblock Elite: Encoding visual concepts into textual embeddings for customized text-to-image generation.
\newblock In \emph{Proceedings of the IEEE/CVF International Conference on Computer Vision}, pages 15943--15953, 2023.

\bibitem[Wei et~al.(2024)Wei, Su, Qin, and Wang]{wei2024mm}
Zhichao Wei, Qingkun Su, Long Qin, and Weizhi Wang.
\newblock Mm-diff: High-fidelity image personalization via multi-modal condition integration.
\newblock \emph{arXiv preprint arXiv:2403.15059}, 2024.

\bibitem[Xiao et~al.(2024)Xiao, Yin, Freeman, Durand, and Han]{xiao2024fastcomposer}
Guangxuan Xiao, Tianwei Yin, William~T Freeman, Fr{\'e}do Durand, and Song Han.
\newblock Fastcomposer: Tuning-free multi-subject image generation with localized attention.
\newblock \emph{International Journal of Computer Vision}, pages 1--20, 2024.

\bibitem[Xie et~al.(2023)Xie, Huang, Dong, Zhao, Dong, Zhang, Zhu, and Liang]{xie2023gp}
Zhenyu Xie, Zaiyu Huang, Xin Dong, Fuwei Zhao, Haoye Dong, Xijin Zhang, Feida Zhu, and Xiaodan Liang.
\newblock Gp-vton: Towards general purpose virtual try-on via collaborative local-flow global-parsing learning.
\newblock In \emph{Proceedings of the IEEE/CVF Conference on Computer Vision and Pattern Recognition}, pages 23550--23559, 2023.

\bibitem[Xu et~al.(2024)Xu, Gu, Chen, and Chen]{xu2403ootdiffusion}
Y Xu, T Gu, W Chen, and C Chen.
\newblock Ootdiffusion: Outfitting fusion based latent diffusion for controllable virtual try-on. arxiv 2024.
\newblock \emph{arXiv preprint arXiv:2403.01779}, 2024.

\bibitem[Ye et~al.(2023)Ye, Zhang, Liu, Han, and Yang]{ye2023ip}
Hu Ye, Jun Zhang, Sibo Liu, Xiao Han, and Wei Yang.
\newblock Ip-adapter: Text compatible image prompt adapter for text-to-image diffusion models.
\newblock \emph{arXiv preprint arXiv:2308.06721}, 2023.

\bibitem[Zhang et~al.(2023{\natexlab{a}})Zhang, Rao, and Agrawala]{zhang2023adding}
Lvmin Zhang, Anyi Rao, and Maneesh Agrawala.
\newblock Adding conditional control to text-to-image diffusion models.
\newblock In \emph{Proceedings of the IEEE/CVF International Conference on Computer Vision}, pages 3836--3847, 2023{\natexlab{a}}.

\bibitem[Zhang et~al.(2024)Zhang, Song, Liu, Wang, Yu, Tang, Li, Tang, Hu, Pan, et~al.]{zhang2024ssr}
Yuxuan Zhang, Yiren Song, Jiaming Liu, Rui Wang, Jinpeng Yu, Hao Tang, Huaxia Li, Xu Tang, Yao Hu, Han Pan, et~al.
\newblock Ssr-encoder: Encoding selective subject representation for subject-driven generation.
\newblock In \emph{Proceedings of the IEEE/CVF Conference on Computer Vision and Pattern Recognition}, pages 8069--8078, 2024.

\bibitem[Zhang et~al.(2023{\natexlab{b}})Zhang, Han, Ghosh, Metaxas, and Ren]{zhang2023sine}
Zhixing Zhang, Ligong Han, Arnab Ghosh, Dimitris~N Metaxas, and Jian Ren.
\newblock Sine: Single image editing with text-to-image diffusion models.
\newblock In \emph{Proceedings of the IEEE/CVF Conference on Computer Vision and Pattern Recognition}, pages 6027--6037, 2023{\natexlab{b}}.

\end{thebibliography}
}

\clearpage
\setcounter{page}{1}
\maketitlesupplementary

In the supplementary material, the sections are organized as follows:
\begin{itemize}
    \item We provide more details regarding parameters, datasets and user study in Sec.~\ref{Sec: implementation}.
    \item We further prove the scalability of AnyDressing in Sec.~\ref{sec: scalability}. 
    \item We provide more ablation results in Sec.~\ref{sec: ablation study}. 
    \item We provide more comparisons with baselines, more qualitative results in the wild and more applications in Sec.~\ref{sec: results}. 
\end{itemize}

\section{Implementation Details} \label{Sec: implementation}
\subsection{Detailed Parameters}
In our experiments, we use SOTA large multi-modal model CogVLM~\cite{wang2023cogvlm} to caption the image. GarmentsNet requires only one step forward process before the multiple denoising steps in DressingNet, causing a minimal amount of extra computational cost. The hyper-parameters used in our experiments are set as follows: 
\begin{itemize}
    \item For the Dressing-Attention mechanism, we set the hyperparameter $\lambda = 0.7$ during inference to get customized results. 
    \item For the noisy timestep threshold discussed in the Garment-Enhanced Texture Learning (GTL) strategy, we set $\eta = 350$. 
    \item The other hyper-parameters used in the experiment are as follows: $\lambda_1 = 0.01$, $\lambda_2 = 0.001$. 
\end{itemize}

\subsection{Datasets} \label{subsec: Datasets} 
\begin{figure*}[t]
  \centering
  \includegraphics[width=0.95\linewidth]{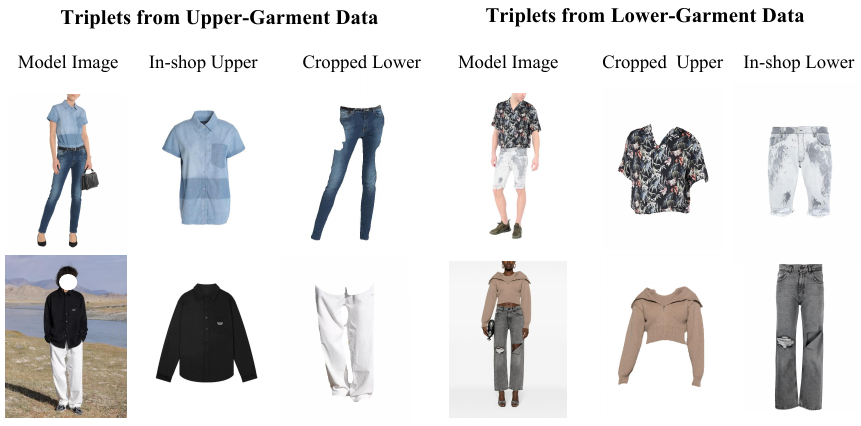}
  \caption{
    Examples of the \textbf{training dataset I}.
  }\label{fig:supp_data1}
\end{figure*}

To facilitate research on multi-garment virtual dressing, a dataset consisting of image triplets is necessary, with each triplet containing an upper garment image, a lower garment image, and a model image wearing the corresponding garments. However, existing in-shop garment to model pairs~\cite{choi2021viton, morelli2022dress} only contain a single reference garment. We leverage the public DressCode dataset along with a proprietary dataset to construct triplets, as illustrated in Fig.~\ref{fig:supp_data1}. 
Assuming we begin with the upper garment data, where we already have an in-shop upper garment and a model image wearing it, we employ human parsing techniques~\cite{jin2023sssegmenation, jin2024idrnet} to roughly segment and extract the lower garment portion from the model image, using it as the corresponding lower garment image. At this stage, the triplet comprises an in-shop upper garment image, a cropped lower garment image, and a model image. Similarly, triplets derived from the lower garment data consist of a cropped upper garment image, an in-shop lower garment image, and a model image. Finally, we constructed 26,114 public triplets from Dresscode and 37,065 triplets from the proprietary dataset to train AnyDressing. 

It is worth noting that our model has not encountered garment pairs in the form of (in-shop upper garment, in-shop lower garment) or (cropped upper garment, cropped lower garment) during training. Nevertheless, it exhibits strong robustness during inference, indicating that the model has effectively learned the proper way to combine upper and lower garments through training.

\subsection{User Study} 
\begin{figure*}[t]
  \centering
  \includegraphics[width=0.95\linewidth]{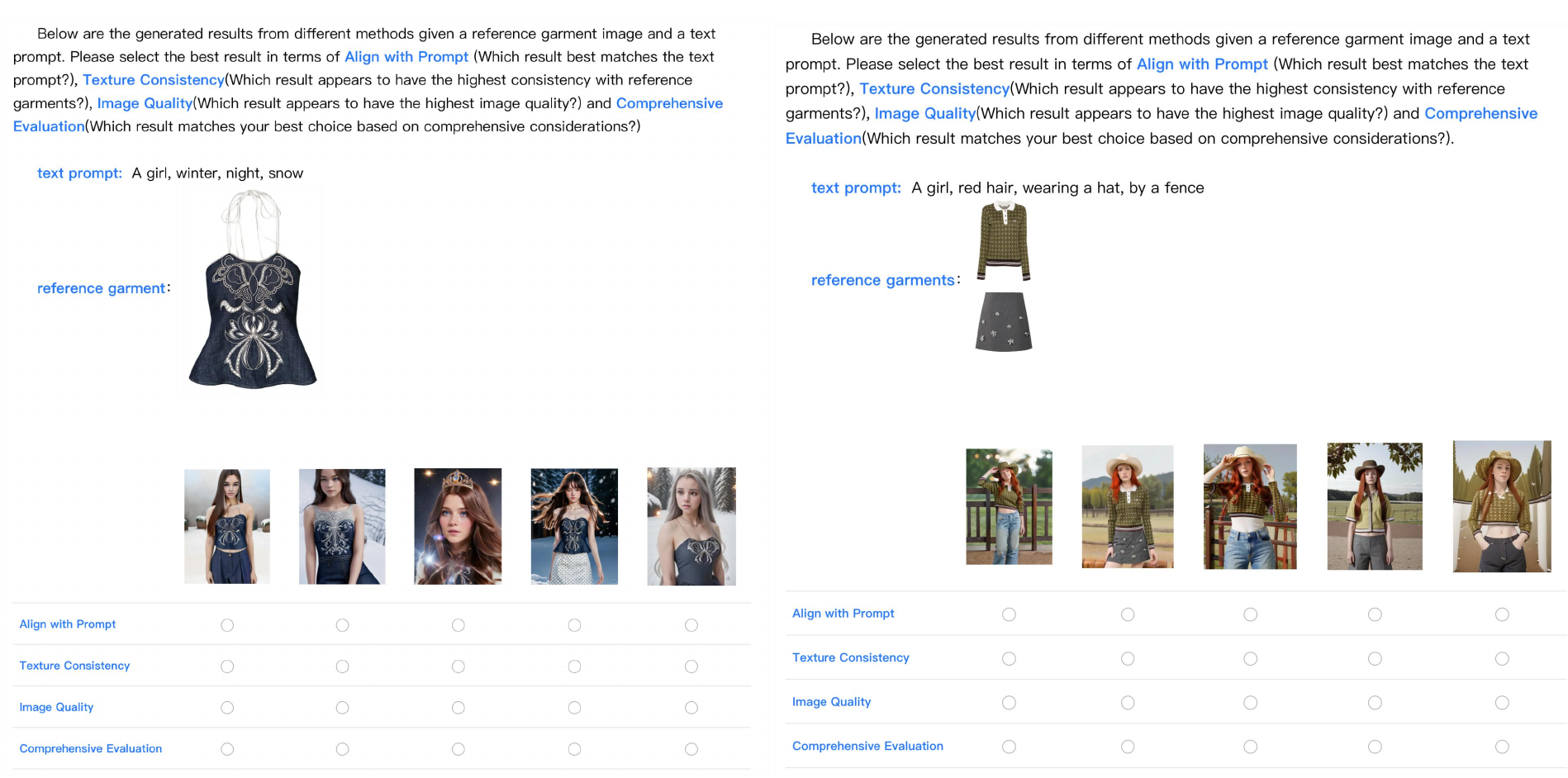}
  \caption{
    Screenshot of \textbf{user study}.
  }\label{fig:user_study}
\end{figure*}

To compare with the baseline methods, we conduct a user study as part of the evaluation. The survey randomly presented 50 sets of generated results to each participant. A screenshot of the survey for a set of generated results is displayed in Fig.~\ref{fig:user_study}, which includes five images and four questions:
\begin{enumerate}
    \item \textit{Which result appears to have the highest consistency with reference garments? }
    \item \textit{Which result best matches the prompt `[prompt]'?}
    \item \textit{Which result appears to have the highest image quality? }
    \item \textit{Which result matches your best choice based on comprehensive considerations? }
\end{enumerate}

For each set of results displayed in the survey, we ensured that their order was randomly shuffled to prevent bias. Responses where all answers had the same selection and responses with completely identical answers were considered invalid. Finally, we obtained a total of 40 valid surveys to evaluate the model.

\section{Scalability of AnyDressing} \label{sec: scalability}
\begin{figure*}[t]
  \centering
  \includegraphics[width=0.95\linewidth]{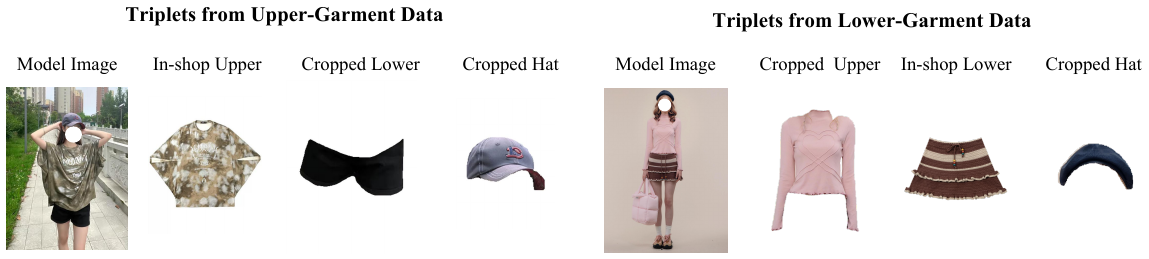}
  \caption{
    Examples of the \textbf{training dataset II}.
  }\label{fig:supp_data2}
\end{figure*}

\begin{figure*}[t]
  \centering
  \includegraphics[width=0.95\linewidth]{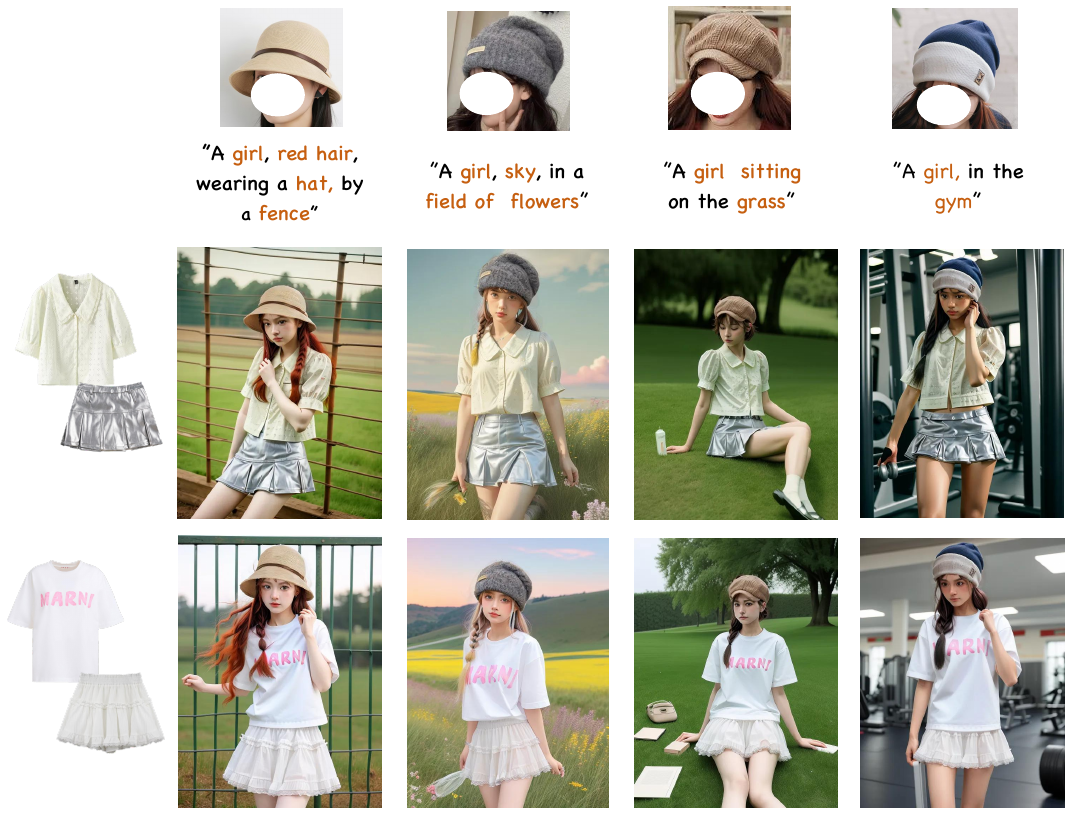}
  \caption{
    Qualitative results of \textbf{more combinations of clothing items}. 
  }\label{fig:hat}
\end{figure*}

To further validate the scalability of our designed GarmentsNet structure, we introduce more combinations of clothing items (hat, upper garment and lower garment), as illustrated in Fig.~\ref{fig:hat}. As shown in Fig.~\ref{fig:supp_data2}, to train the model, we construct datasets using the same idea as introduced in Sec.~\ref{subsec: Datasets}. Specifically, we select 18,059 pairs from the proprietary dataset that satisfies the model image containing the hat, and use the human parsing techniques to obtain the cropped hat image from the model image.

Notably, each additional garment condition requires only some newly added LoRA matrix $\triangle\hat{\mathbf{W}}$ in the Garment-Specific Feature Extractor (GFE) module. And it requires only a single forward pass (timestep $t = 0$) to encode the clothing before injecting features into the DressingNet, minimizing the additional computational time during both the training and inference process. 
This experiment effectively demonstrates that our GarmentsNet can be extended to accommodate any number of clothing items. 
Additionally, thanks to our proposed Instance-Level Garment Localization (IGL) learning mechanism, AnyDressing can further prevent garment blending and enhance fidelity to customized text prompts.

\section{More Ablation Study} \label{sec: ablation study}
\begin{figure*}[t]
  \centering
  \includegraphics[width=0.85\linewidth]{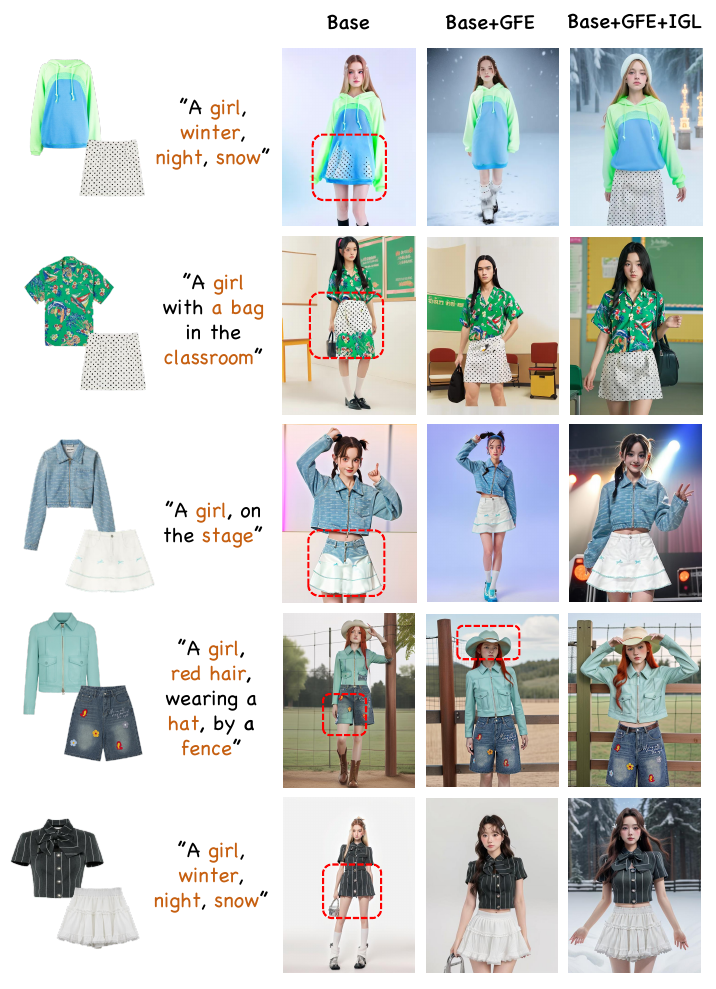}
  \caption{
    More \textbf{ablation results on GFE and IGL modules}.
  }\label{fig:supp_abl}
\end{figure*}

In Fig.~\ref{fig:supp_abl}, we present additional visual results to validate the effectiveness of the Garment-Specific Feature Extractor (GFE) module and the Instance-Level Garment Localization (IGL) learning mechanism. We employ traditional ReferenceNet~\cite{hu2024animate} to encode multiple garments concurrently and then incorporate them into the denoising U-Net similar to~\cite{shen2024imagdressing, chen2024magic} as our base model. 
As shown in Fig.~\ref{fig:supp_abl}, \textbf{Base} model encounters severe clothing confusion issues, resulting in the colors and patterns of multiple garments blending. In contrast, \textbf{Base+GFE} significantly reduces garment confusion and improves garment consistency, which is attributed to the multi-garment parallel processing design of our designed GFE module. \textbf{Base+GFE+IGL} shows better fidelity to the text prompts and further mitigates background contamination, which demonstrates IGL mechanism effectively constrains garment features to attend to the correct regions and avoid influencing other irrelevant regions in the synthetic images. 

\section{More Results} \label{sec: results}
\subsection{More Comparisons}
\begin{figure*}[t]
  \centering
  \includegraphics[width=0.8\linewidth]{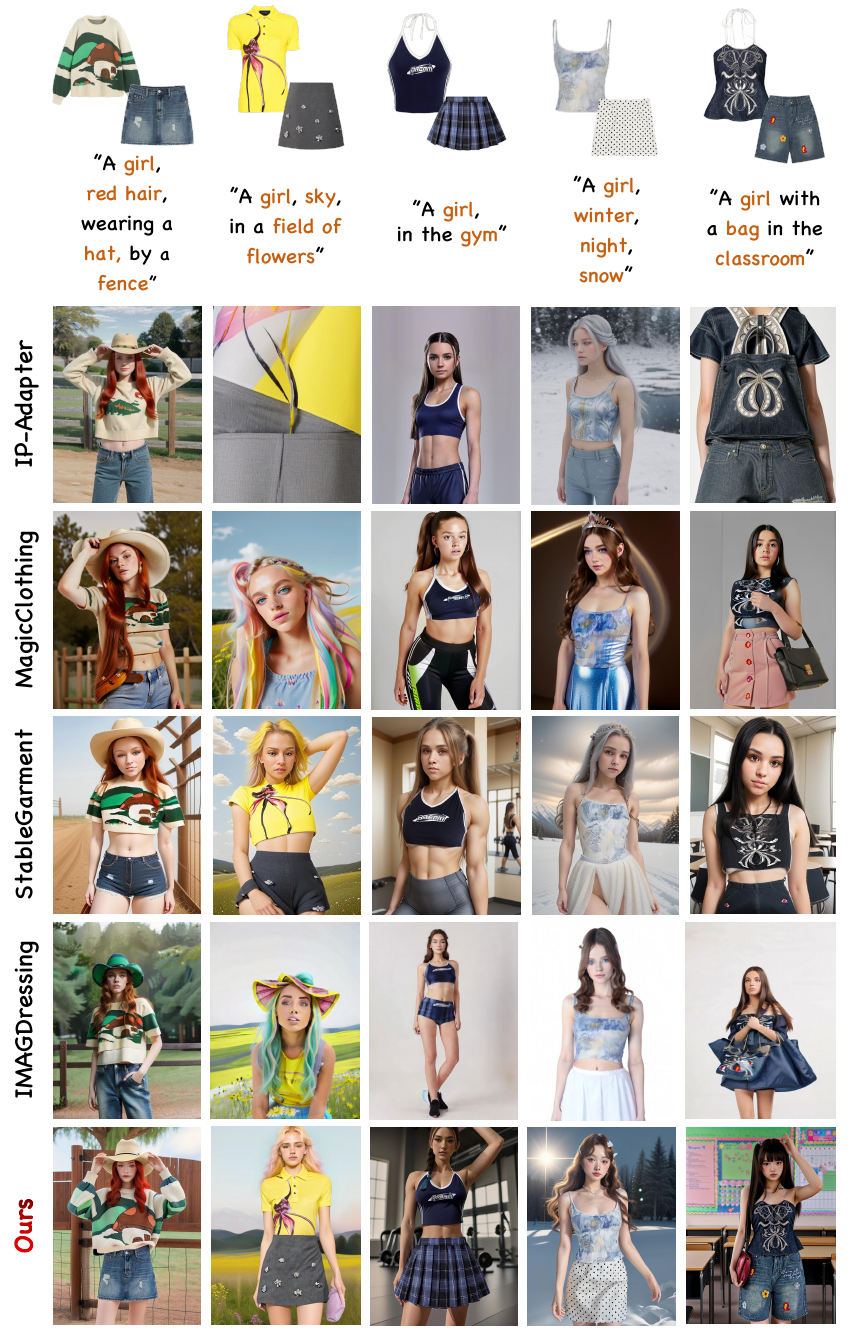}
  \caption{
    More \textbf{qualitative comparisons I}.
  }\label{fig:supp_comp1}
  \vspace{-1.01073pt}
\end{figure*}

\begin{figure*}[t]
  \centering
  \includegraphics[width=0.8\linewidth]{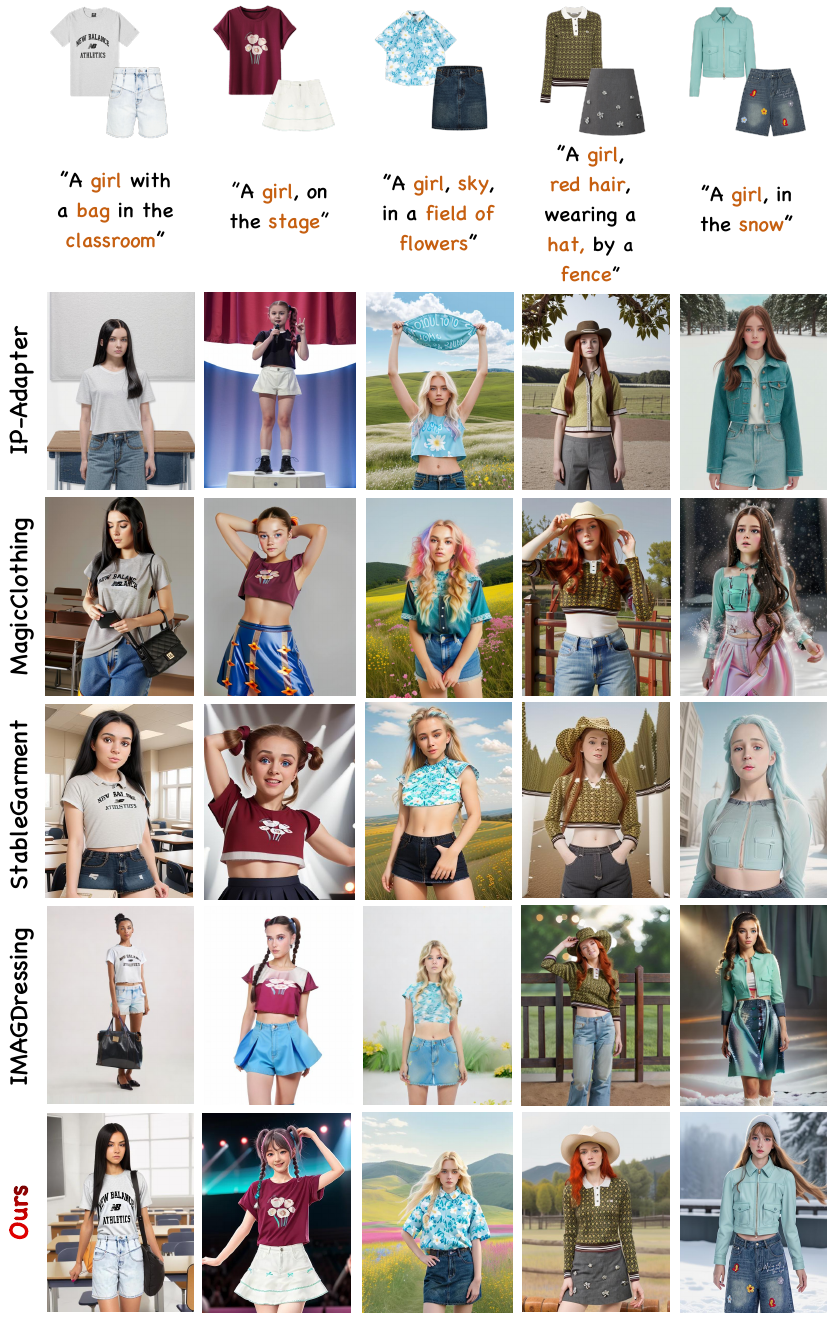}
  \caption{
    More \textbf{qualitative comparisons II}.
  }\label{fig:supp_comp2}
  \vspace{-5.54272pt}
\end{figure*}

As shown in Fig.~\ref{fig:supp_comp1}-\ref{fig:supp_comp2}, We provide more visual comparisons between our method and state-of-the-art baselines~\cite{ye2023ip, chen2024magic, wang2024stablegarment, shen2024imagdressing}. It is clear from these comparisons that our method maintains superior consistency in clothing style and texture, and exhibits better text fidelity. 

\subsection{More Visual Results}
\begin{figure*}[t]
  \centering
  \includegraphics[width=0.9\linewidth]{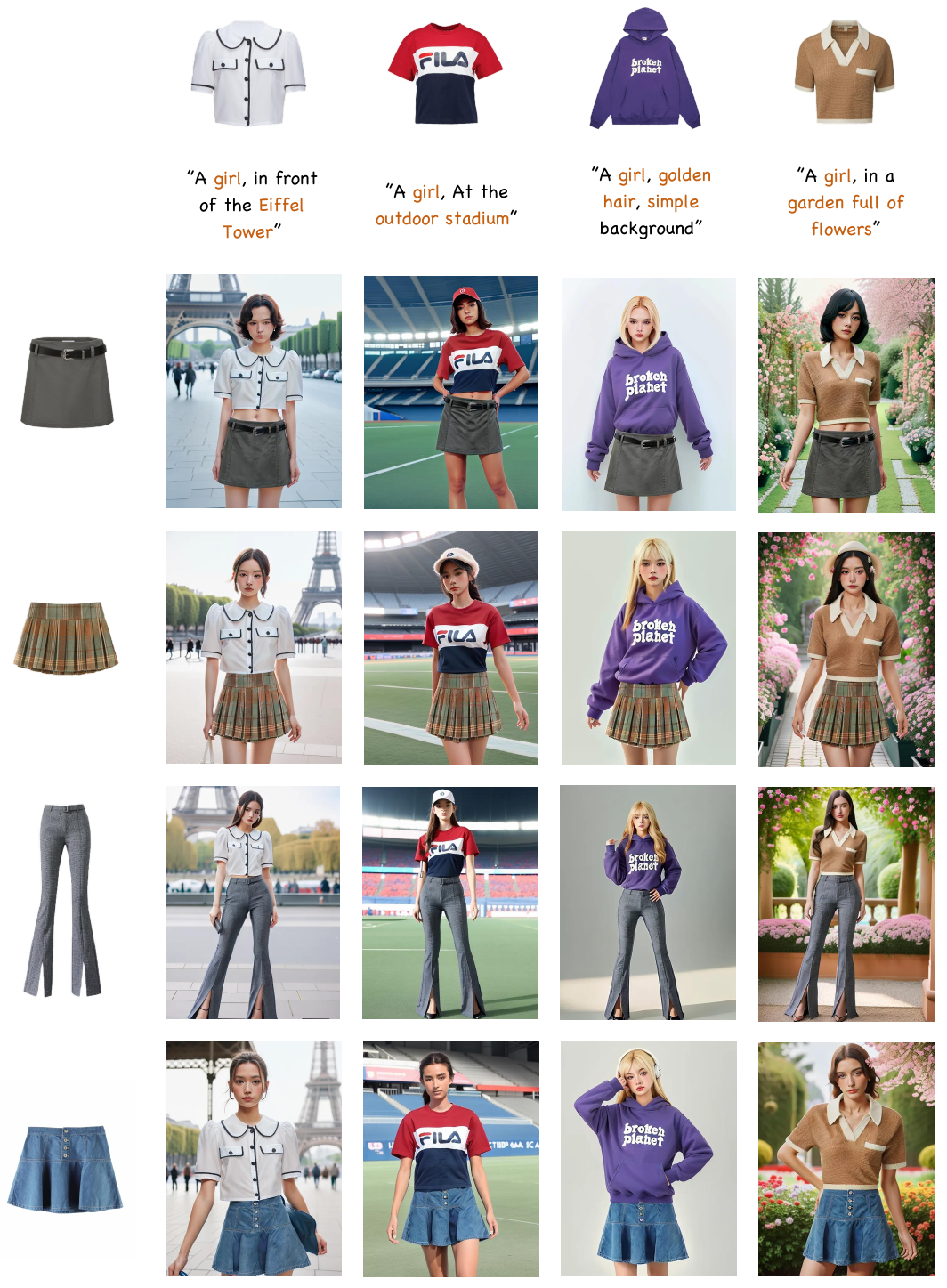}
  \caption{
    More \textbf{qualitative results I}.
  }\label{fig:supp_ours1}
\end{figure*}

\begin{figure*}[t]
  \centering
  \includegraphics[width=0.9\linewidth]{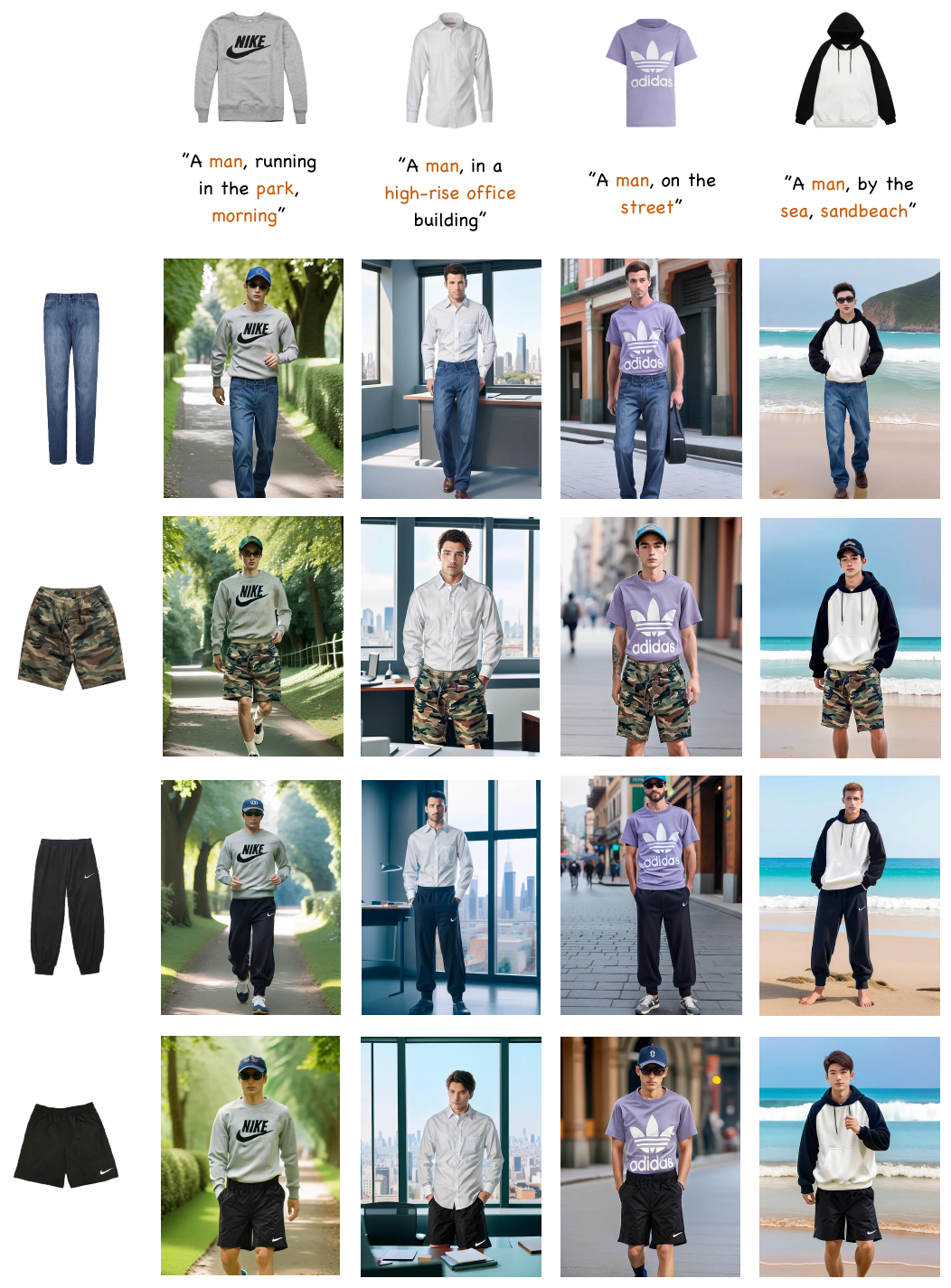}
  \caption{
    More \textbf{qualitative results II}.
  }\label{fig:supp_ours2}
\end{figure*}

As shown in Fig.~\ref{fig:supp_ours1}-\ref{fig:supp_ours3}, we provide more multi-garment virtual dressing results of AnyDressing in the wild. 
It can be observed that our method produces high-quality customized virtual dressing results for various types of garment combinations, while faithfully adhering to personalized text prompts. Experiments in complex scenarios demonstrate that AnyDressing significantly enhances the practical application of Virtual Dressing in e-commerce and creative design.

\subsection{More Applications}
\begin{figure*}[t]
  \centering
  \includegraphics[width=0.95\linewidth]{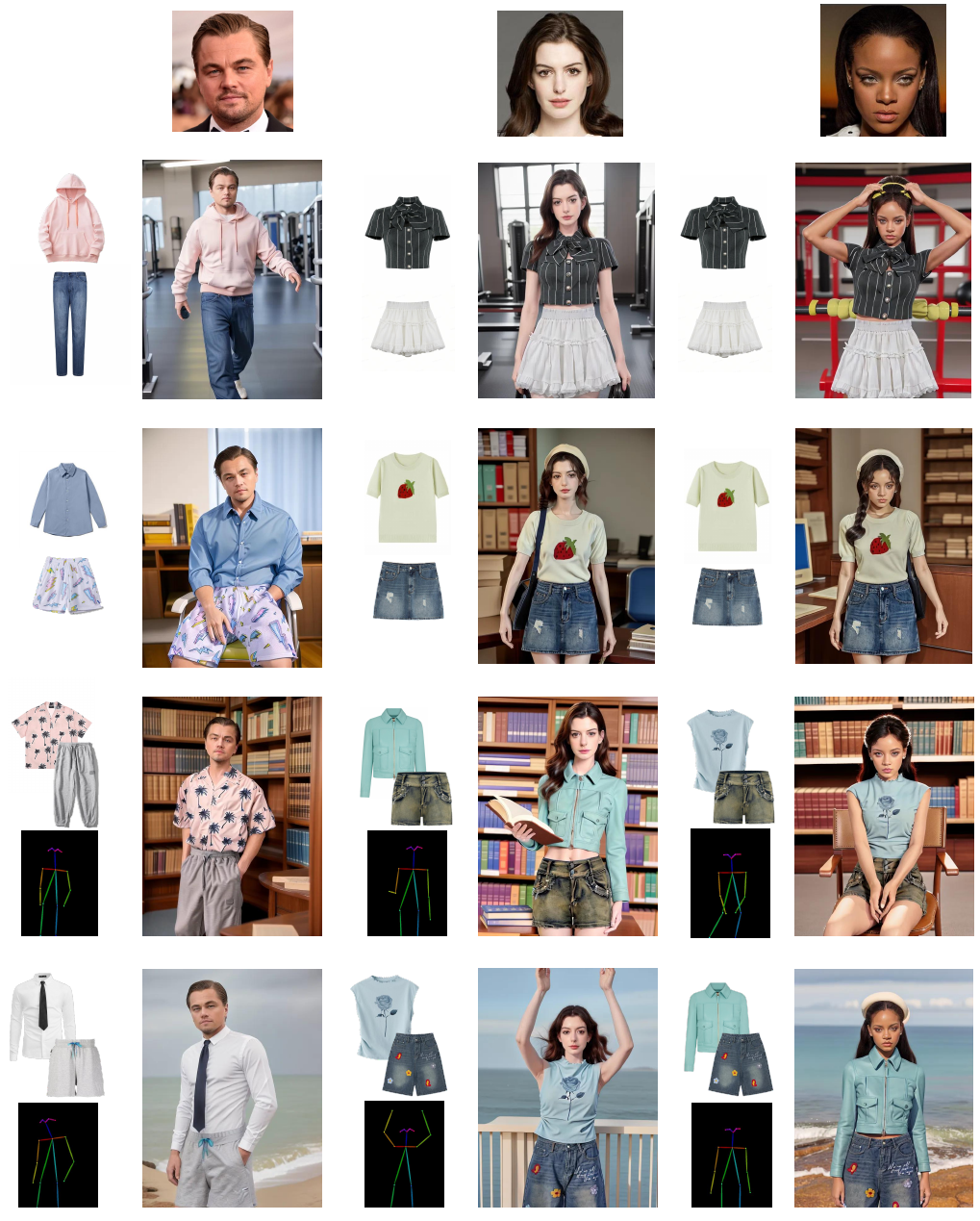}
  \caption{
    More results of \textbf{combining ControlNet~\cite{zhang2023adding} and FaceID~\cite{ye2023ip}}.
  }\label{fig:supp_ipa}
\end{figure*}

\textbf{Combined with ControlNet.} Leveraging the capabilities of ControlNet~\cite{zhang2023adding},
our model can generate personalized models guided by specific conditions. We present the OpenPose-guided generation results in Fig.~\ref{fig:supp_ipa}. 

\noindent \textbf{Combined with IP-Adapter.} Our model enables the generation of target individuals wearing specified garments integrated with the IP-Adapter. We utilize the ID preservation capability of FaceID~\cite{ye2023ip} to provide an authentic virtual dressing experience. The visual results, as shown in Fig.~\ref{fig:supp_ipa}. 

\begin{figure*}[t]
  \centering
  \includegraphics[width=0.9\linewidth]{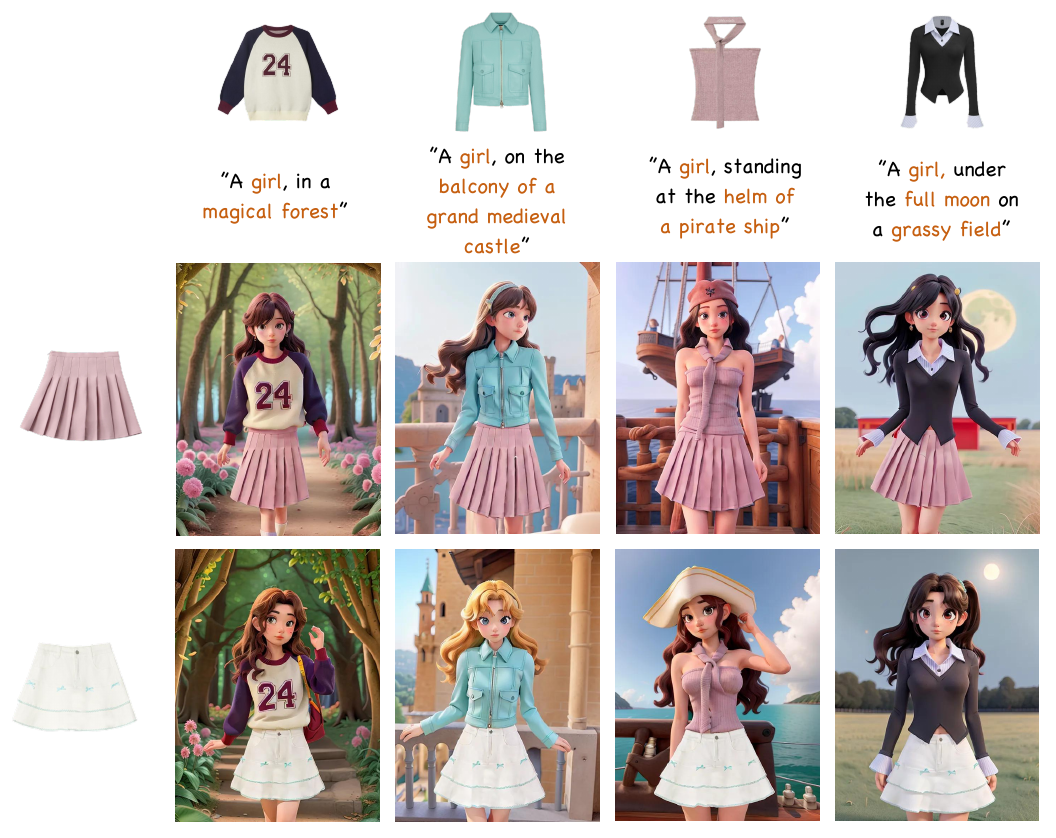}
  \vspace{-3mm}
  \caption{
    More \textbf{qualitative results III}.
  }\label{fig:supp_ours3}
  \vspace{-2mm}
\end{figure*}

\begin{figure*}[t]
  \centering
  \includegraphics[width=0.95\linewidth]{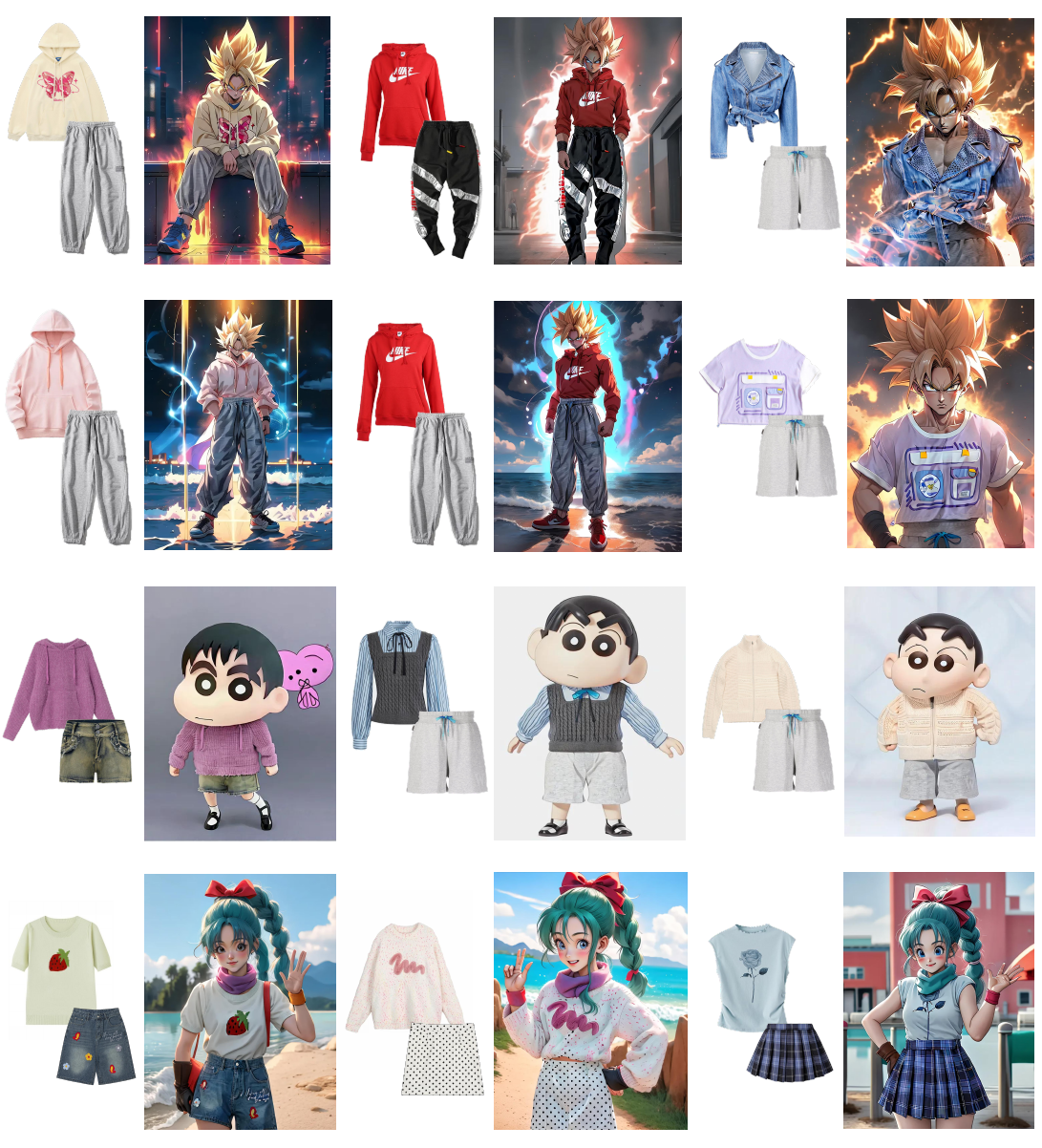}
  \caption{
    More results of \textbf{combining LoRAs~\cite{hu2021lora}}.
  }\label{fig:supp_lora}
\end{figure*}

\noindent \textbf{Stylized Customization.} Furthermore, by utilizing stylized base models or customized LoRAs~\cite{hu2021lora}, we can generate creative and stylized outputs while preserving the intricate details of the garments, as shown in Fig.~\ref{fig:supp_ours3} and Fig.~\ref{fig:supp_lora}.






\end{document}